%% file: main.tex
\documentclass[10pt,twocolumn,letterpaper]{article}

\PassOptionsToPackage{table}{xcolor}
\usepackage{iccv}              

\usepackage{multirow}
\usepackage[title]{appendix}
\usepackage[accsupp]{axessibility}
\usepackage{ulem}
\usepackage{graphicx}
\usepackage{fontawesome}
\usepackage{makecell}
\usepackage{amssymb}
\usepackage{pifont}

\usepackage{tcolorbox}          
\usepackage{listings}           
\tcbuselibrary{listings}        
\tcbuselibrary{breakable}       

\input{preamble}

\definecolor{iccvblue}{rgb}{0.21,0.49,0.74}
\usepackage[pagebackref,breaklinks,colorlinks,allcolors=iccvblue]{hyperref}

\def\paperID{1660} 
\def\confName{ICCV}
\def\confYear{2025}

\title{RoboFactory: Exploring Embodied Agent Collaboration \\ with Compositional Constraints}

\author{
Yiran Qin\textsuperscript{1,2}\footnotemark[1], \quad
Li Kang\textsuperscript{2,6}\footnotemark[1], \quad
Xiufeng Song\textsuperscript{2,7}\footnotemark[1], \quad
Zhenfei Yin\textsuperscript{3}\footnotemark[2], \\
Xiaohong Liu\textsuperscript{7}, \quad
Xihui Liu\textsuperscript{4}, \quad
Ruimao Zhang\textsuperscript{5}\footnotemark[2], \quad
Lei Bai\textsuperscript{2}\footnotemark[2]\\
\small$^{1}$The Chinese University of Hong Kong, Shenzhen ~~
\small$^{2}$Shanghai Artificial Intelligence Laboratory~~\\
\small$^{3}$Oxford ~~
\small$^{4}$HKU~~
\small$^{5}$Sun Yat-sen University~~
\small$^{6}$Tongji University~~
\small$^{7}$Shanghai Jiao Tong University\\
\tt\footnotesize yiranqin@link.cuhk.edu.cn~~~faceong02@gmail.com~~~akikaze@sjtu.edu.cn~~~\\
\url{https://iranqin.github.io/robofactory/}
}

\begin{document}

\twocolumn[{
\renewcommand\twocolumn[1][]{#1}
\maketitle
\centering
\vspace{-3mm}
\includegraphics[width=\textwidth]{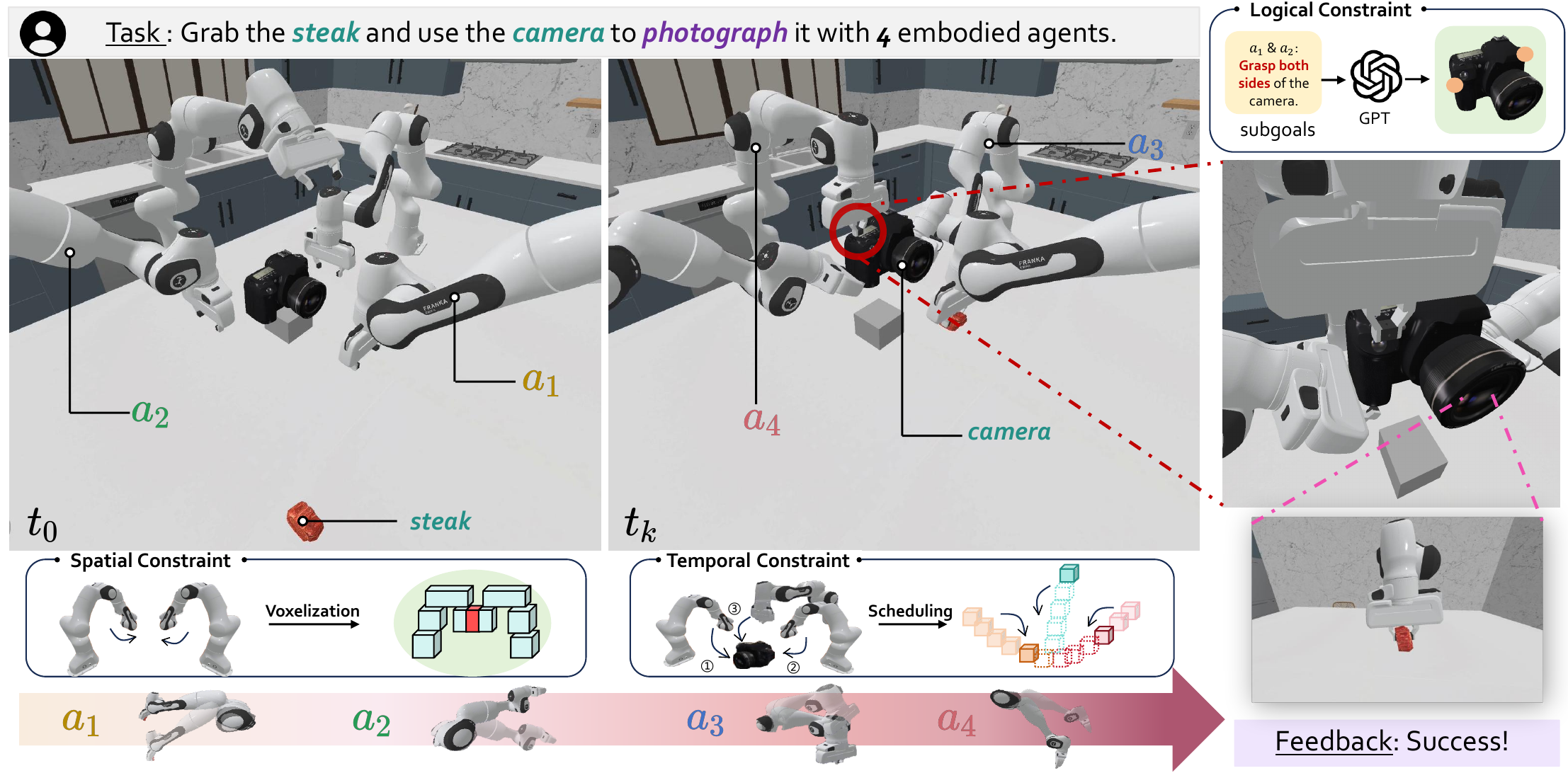}
\captionsetup{type=figure}
\vspace{-3mm}
\caption{When performing the task ``Grab the steak and use the camera to photograph it with 4 embodied agents", collaboration among multiple agents is required: $a_1$ grasps the steak, $a_2$ and $a_3$ lift the camera, and $a_4$ presses the shutter to take the photo. However, each agent cannot focus solely on its own task. We introduce the concept of compositional constraints to ensure safe and efficient collaboration among the agents. Logical constraints prevent incorrect interaction forms (\eg, $a_3$ grabbing the camera lens, causing damage). Spatial constraints avoid catastrophic hardware damage (\eg, collisions between $a_2$ and $a_3$ during trajectory execution). Temporal constraints prevent inefficient collaboration (\eg, $a_1$ waiting unnecessarily due to nonexistent collisions while other agents execute their tasks).}
\label{fig:overview}
\vspace{5mm}
}]

\let\thefootnote\relax\footnotetext{$^*$ Equal contribution\hspace{3pt} \hspace{5pt}$^\dagger$ Corresponding author\hspace{5pt}
}

\input{sec/0_abstract}

\input{sec/1_intro}
\input{sec/2_relatedwork}
\input{sec/3_method}
\input{sec/4_exp}
\input{sec/5_conclusion}
{
    \small
    \bibliographystyle{ieeenat_fullname}
    \bibliography{main}
}

\input{sec/6_sup}

\end{document}

%% file: preamble.tex
\newtcblisting{promptcode}[1][Python Prompt Example]{
    listing only,
    listing options={
        breaklines=true,       
        breakatwhitespace=true,
        keepspaces=true,       
        basicstyle=\ttfamily\scriptsize, 
        showstringspaces=false,
        tabsize=4,            
        columns=flexible,      
        escapeinside={/*}{*/},  
    },
    breakable=true,           
    width=\linewidth,         
}

\newtcblisting{pythoncode}[1][Python Prompt Example]{
    listing only,
    listing options={
        language=Python,
        breaklines=true,       
        breakatwhitespace=true,
        keepspaces=true,       
        basicstyle=\ttfamily\scriptsize, 
        showstringspaces=false,
        tabsize=4,            
        columns=flexible,      
        escapeinside={/*}{*/},  
    },
    breakable=true,           
    width=\linewidth,         
}

\newcommand{\cmark}{\ding{51}}
\newcommand{\xmark}{\ding{55}}

\newcommand{\mname}{\textsl{RoboFactory}}

%% file: sec/0_abstract.tex
\begin{abstract}
Designing effective embodied multi-agent systems is critical for solving complex real-world tasks across domains.
Due to the complexity of multi-agent embodied systems, existing methods fail to automatically generate safe and efficient training data for such systems.
To this end, we propose the concept of compositional constraints for embodied multi-agent systems, addressing the challenges arising from collaboration among embodied agents.
We design various interfaces tailored to different types of constraints, enabling seamless interaction with the physical world.
Leveraging compositional constraints and specifically designed interfaces, we develop an automated data collection framework for embodied multi-agent systems and introduce the first benchmark for embodied multi-agent manipulation, \mname{}.
Based on \mname{} benchmark, we adapt and evaluate the method of imitation learning and analyzed its performance in different difficulty agent tasks.
Furthermore, we explore the architectures and training strategies for multi-agent imitation learning, aiming to build safe and efficient embodied multi-agent systems.
Please see the project page at \url{https://iranqin.github.io/robofactory/}.

\end{abstract}

%% file: sec/1_intro.tex
\section{Introduction}
\label{sec:intro}

With the increasing diversity of robotic forms and the advancement of robotic control strategies, current robotic systems are capable of performing fixed tasks~\cite{zhao2023learning,chi2023diffusion} or executing tasks based on instructions~\cite{black2024pi_0,kim2024openvla}, offering infinite possibilities to build interactive agents in the real world.
While these robotic systems typically focus on single-agent tasks, many real-world applications—such as manufacturing and medical assistance—require multiple embodied agents to collaborate on tasks. 
Such collaboration is vital for tackling tasks that exceed the capabilities of a single agent and enhancing task efficiency through multi-agent deployment.
%


To train the policies of multi-embodied agents, researchers often need to collect data through simultaneous remote operation by multiple people, which is extremely inefficient.
%
With the rise of large language models (LLMs)~\cite{achiam2023gpt,touvron2023llama}, many single-agent works have attempted to leverage their powerful reasoning capabilities to automate the data generation process~\cite{mu2024robotwin,mu2024robocodex}, using predefined motion primitives to carry out physical interactions with the environment. 
This significantly reduces the labor and time costs associated with data collection.
However, when it comes to the more challenging tasks of multi-agent collaboration, the dimensions that need to be considered become far more complex:
1) Plan global tasks and allocate agents based on task logic and agent availability, ensuring logical consistency and maximizing efficiency.
2) Manage shared physical space to prevent collisions between embodied agents.
3) Schedule agents sharing the same space at different timesteps or perform some low-level operations simultaneously to improve system efficiency. 
%
Simple adaptations of single-agent embodied systems cannot meet such requirements, and additional constraints need to be introduced.

%
We first introduce \textit{compositional constraints} based on three specific constraint for multi-agent embodied data generation.
%
Each constraint is specifically designed for multi-agent collaborative tasks and restricts the behavior of embodied agents in the corresponding dimension (\eg, space).
%
By combining these constraints, agents could conduct more reasonable and safe behaviors that aligns with real-world embodied agent collaboration scenarios:
1) \textbf{Logical Constraints}: Define the rules and relationships that govern valid robot behaviors.
2) \textbf{Spatial Constraints}: Specify constraints based on physical or spatial boundaries.
3) \textbf{Temporal Constraints}: Impose constraints related to time to make more efficient collaboration.

In this work, we propose a framework \mname{} for multi-agent embodied data generation based on \textit{compositional constraints}, which not only benefits from the generalization power of LLMs but also satisfies the additional requirements of multi-agent tasks.
%
%
Given the global task description, prior information, and observations, RoboBrain generates the next sub-goals for each agent, outputs textual compositional constraints, and produces unconstrained trajectory sequences for each agent by invoking predefined motion primitives to achieve these sub-goals.
However, the compositional constraints generated by RoboBrain are limited to the textual modality, making them insufficient for directly constraining real-world decision-making.
To address this limitation, we designed various interfaces tailored to different types of constraints, enabling seamless interaction with the physical world.
As a result, RoboChecker can construct corresponding constraint interfaces based on these textual constraints and the current multi-agent state to ensure agents do not violate any constraints while executing the generated trajectories.
%
%
%

Within this framework, \mname{} addresses the challenges of scaling up from single-agent embodied systems to multi-agent embodied systems by constructing a safe and efficient data production pipeline. Using RoboFactory, we designed various multi-agent collaboration scenarios and evaluated the design of embodied multi-agent systems within these settings.
First, we validated the necessity of compositional constraints in multi-agent data generation. We then deployed the diffusion policy on several multi-agent collaborative tasks and conducted extensive testing. Furthermore, we explored the design of multi-agent systems based on imitation learning, investigating more effective architectural designs. 
Our contributions can be summarized as follows:

\begin{enumerate}
    \item We propose the concept of compositional constraints for embodied multi-agent systems, addressing the challenges arising from collaboration among embodied agents.
    \item Leveraging compositional constraints and specifically designed interfaces, we develop an automated data collection framework for embodied multi-agent systems and introduce the first benchmark for embodied multi-agent manipulation, \mname{}.
    \item Based on \mname{}, we deploy imitation learning methods and conduct evaluations, and explore the architectures and training strategies for multi-agent imitation learning, aiming to build safe and efficient embodied multi-agent systems.
\end{enumerate}

%% file: sec/2_relatedwork.tex
\section{Related Work}
\label{sec:related_word}

\subsection{Multi-Agent System}
Multi-agent systems consist of multiple autonomous entities, each with access to distinct information and potentially conflicting objectives. 
Based on their functionalities, recent multi-agent systems can generally be categorized into two types: tool-based agent assistants~\cite{zhao2024expel,mosquera2024can,wang2024sibyl,li2023camel,wu2023autogen,hong2023metagpt,yang2024oasis} and simulation environments for societies or games~\cite{zhou2023webarena,huang2024far,yu2024mineland,qin2024mp5}.
Different from the above work, \mname{} focuses more on the application of multi-agent collaboration in real-world decision-making, especially the low-level manipulation of embodied agents.

\subsection{Robot Manipulation}
Behavioral Cloning (BC)~\cite{Dalal2023ImitatingTA,jang2022bc,jiang2023vima,mandlekar2020learning,ma2024contrastive} trains policies using pre-recorded human demonstrations to directly imitate expert behaviors, whereas Offline Reinforcement Learning (ORL)~\cite{kalashnikov2021mt,kumar2022pre,chebotar2023q} refines action selection through reward maximization across extensive datasets. Generative approaches have expanded methodology: Action Chunking with Transformers (ACT) employs Transformer architectures combined with conditional variational autoencoders to model sequential decision-making~\cite{vaswani2017attention,zhao2023learning,buamanee2024bi}. Meanwhile, diffusion-based frameworks have gained traction in robotic imitation learning for their superior generation performance. Notable examples include Diffusion Policy~\cite{chi2023diffusion} and its 3D variant~\cite{Ze2024DP3} that utilizes point cloud observations to improve geometric understanding. 
Demonstration acquisition primarily relies on human-operated robotic systems across diverse tasks~\cite{ebert2021bridge,mandlekar2020learning,mandlekar2018roboturk,jang2022bc}, with simulator-based trajectory synthesis providing supplementary data sources~\cite{mu2024robotwin,james2020rlbench,taomaniskill3,nambiar2024automation,soroush2024robocasa,qin2024worldsimbench}. Although these robotic systems demonstrate the ability to generate data, they have rarely explored effectively multi-agent robotic manipulation.

\subsection{Visual Programming}
Executing visual programming necessitates robust comprehension of visual concepts and spatio-temporal reasoning capabilities. While existing approaches demonstrate broad applicability in zero-shot scenario~\cite{Cho2023VPT2I,gupta2023visual,10500490,liang2023code,zhou2024minedreamer,qin2025navigatediff} through fusion of large language models (LLMs) with vision systems, they often sacrifice granular precision in task execution. Contemporary methods~\cite{venuto2024code,mu2024robocodex,huang2024story3d} investigate vision-language models (VLMs) for visual programming. CaM~\cite{zhou2024code} introduces constraint-based elements for program synthesis. In our work, \mname{} generates robot trajectories through action primitives, further regulating the effectiveness of visual programming via compositional constraint interface.

%% file: sec/3_method.tex
\section{Compositional Constraints}
\label{sec:CC}
\begin{figure*}[ht]
  \centering
  \includegraphics[width=1\linewidth]{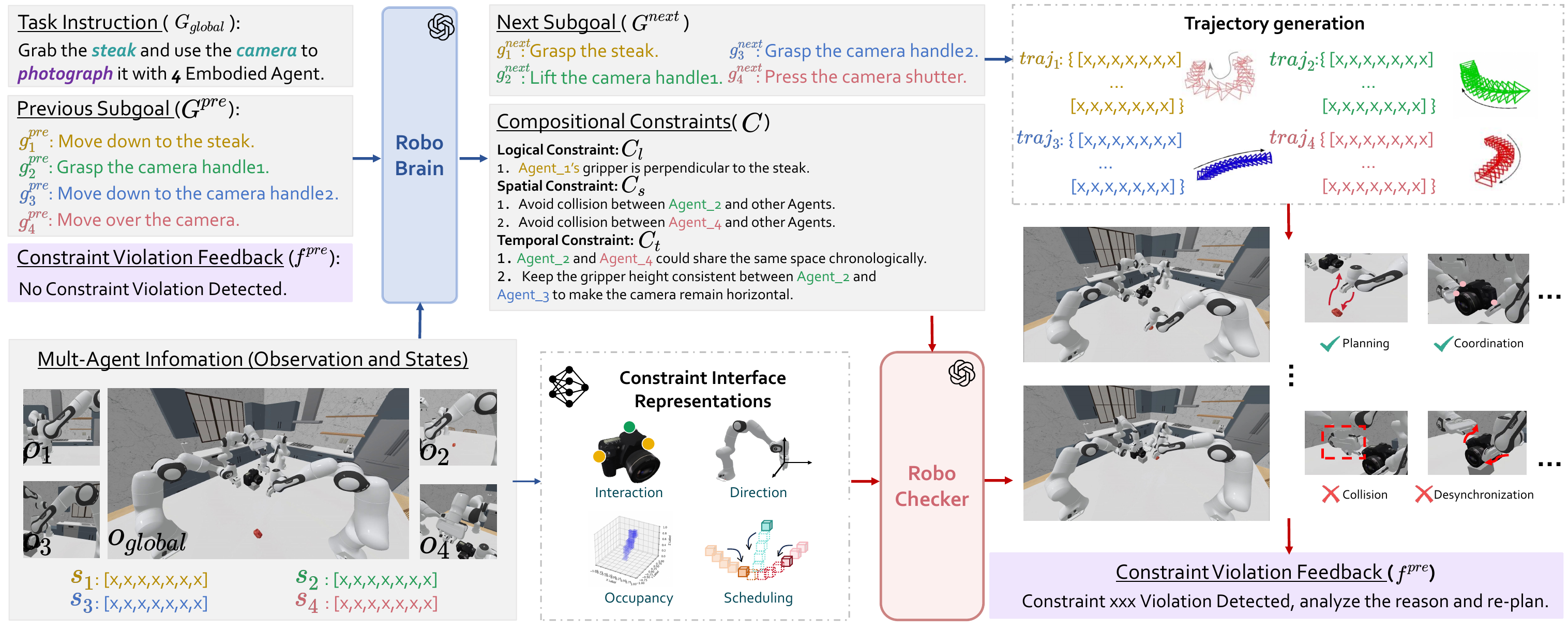}
  \caption{Overview of \mname{}. Given the global task description, prior information, and observations, RoboBrain generates the next sub-goals for each agent and outputs textual compositional constraints. It then generates unconstrained trajectory sequences for each agent to achieve the corresponding sub-goals, invoking predefined motion primitives. RoboChecker constructs corresponding constraint interfaces based on the textual compositional constraints and the current multi-agent state. It checks whether the agents violate any constraints while executing the generated trajectories. This framework ensures the generation of safe and efficient collaborative data for multi-embodied agents by transforming abstract textual constraints into representations that can interact with agent behaviors through the construction of constraint interfaces.}
\label{fig:framework} 
\end{figure*}
Constraints define practical and efficient boundaries grounded in real-world conditions.
In multi-robot collaboration scenarios, the more complex decision space requires various constraints to ensure safety and effectiveness. 
Below, we propose three categories of common constraints that are crucial for modeling real-world boundaries in multi-embodied agent decision-making.
\vspace{-2mm}
\paragraph{Logical Constraints.} Logical Constraints define the permissible actions and interaction rules for agents, focusing on high-level logic such as interaction objects, contact points, and movement directions rather than the timing or sequence of operations. These constraints encode structural rules within task scenarios, such as usage permissions for interactive objects (\eg, only specific tools can be used for processing certain materials), contact point restrictions (\eg, agents must grasp objects from designated points), and directional consistency (\eg, when multiple agents transport an object, their applied forces must remain aligned). By formalizing interaction relationships—such as action compatibility, interaction constraints, and spatial coordination requirements—logical constraints ensure agents follow coherent operational procedures.
\vspace{-2mm}
\paragraph{Spatial Constraints.} Spatial Constraints defines where agents can operate and how physical interactions are structured. These include geometric boundaries (\eg, no agent may enter a 1-meter radius around active machinery), collaborative workspace partitioning (\eg, dividing a construction site into exclusive zones to prevent collisions), and task-specific placement requirements (\eg, components must be positioned within 2 cm of their target coordinates for valid assembly). They also govern adaptive spatial behaviors, such as dynamic rerouting around newly detected obstacles or adjusting gripper orientations to fit narrow apertures. By isolating physical feasibility from temporal and logical considerations, these constraints ensure agents operate within safe, structurally coherent environments.
\vspace{-2mm}
\paragraph{Temporal Constraints.} Temporal Constraints regulate when and in what order actions must be executed, addressing synchronization, deadlines, and sequential dependencies. These constraints ensure that agents align their behaviors with time-sensitive requirements, such as enforcing phased workflows (\eg, Agent C must wait 5 seconds after Agent D finishes welding to begin inspection) or coordinating parallel actions with strict time windows (\eg, two agents must lift an object simultaneously within a 0.5-second tolerance). They also manage dynamic adjustments, such as extending task durations in response to environmental delays or rescheduling actions when prior steps overrun. Unlike logical constraints, which define decision validity, temporal constraints focus strictly on timing feasibility—ensuring actions occur neither prematurely nor too late to maintain safety and efficiency.
\vspace{-2mm}
\paragraph{Compositional Constraints.} The collaboration of multi-embodied agents relies on the integration of logical, temporal, and spatial constraints. 
Logical constraints define interaction protocols and shared objectives, temporal constraints synchronize actions with task dependencies, and spatial constraints encode geometric and semantic boundaries. 
Together, these constraints balance decentralized autonomy with global coherence, enabling conflict resolution, resource optimization, and adaptability. 
This unified framework ensures that local decisions converge into robust, efficient, and executable collaborative behaviors.

\begin{figure*}[ht]
  \centering
  \includegraphics[width=1\linewidth]{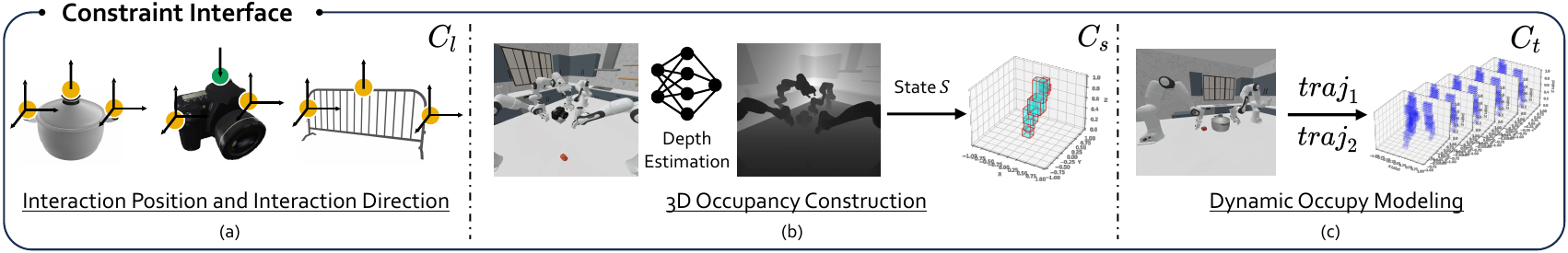}
  \caption{Different Constraint Interface. For $C_l$, we annotated the interactive points of objects and the interactive directions of each point. For $C_s$, we modeled observations to obtain depth maps and used them, along with the robotic arm states, to construct 3D occupancy representations. For $C_t$, we modeled temporal-state representations based on the trajectories of agents at each changing position and used these representations for scheduling through analysis.
  }
\label{fig:constraint} 
\end{figure*}

\section{RoboFactory}
\label{sec:method}
We first give an overview of the proposed \mname{} (Sec.~\ref{sec:method_overview}). 
Then, we elaborate on the Constraint Interface Generation (Sec.~\ref{sec:method_constraint_interface}).
Finally, we present the Dataset and Benchmark of \mname{} (Sec.~\ref{sec:method_benchmark}).

\subsection{Overview}
\label{sec:method_overview}
The proposed RoboFactory framework consists of two core modules, RoboBrain and RoboChecker. 
%
Our work focuses on long-horizon multi-agent manipulation task instructions $\mathcal{G}_{\texttt{global}}$ (\eg, ``Grab the steak and use the camera to photograph it with 4 Embodied Agents."), utilizing RGB observations $\mathcal{O} = \{o_{\texttt{global}}, o_1, ..., o_n\}$, which consist of one global view and multiple ego-centric views from $n$ agents $\{a_1,a_2,..a_n\}$ (\eg, $n = 4$ in this case).
As illustrated in Figure~\ref{fig:framework}, the RGB images $\mathcal{O}$, combined with the text instructions $\mathcal{G}_{\texttt{global}}$, the previous subgoal sets $\mathcal{G}^{\texttt{pre}} = \{g_1^{\texttt{pre}}, ..., g_n^{\texttt{pre}}\}$, and the Constraint Violation Feedback (\eg, subgoal collaboration success or constraint violation details) from RoboChecker, denoted as $f^{\texttt{pre}}$, are input into the RoboBrain $\mathcal{F}_{\texttt{VLM}}$ (\eg, GPT-4o~\cite{achiam2023gpt}).
The model then generates the next subgoal sets $\mathcal{G}^{\texttt{next}} = \{g_1^{\texttt{next}}, ..., g_n^{\texttt{next}}\}$ along with the corresponding textual compositional constraints $\mathcal{C} = \{\mathcal{C}_l, \mathcal{C}_s, \mathcal{C}_t\}$, each type of constraint set contains multiple specific constraints $c$ (\eg, ``Avoid collision between $a_2$ and other agents.").
Here, $\mathcal{C}_l$ represents the logical constraint set that multi-agents must adhere to based on prior knowledge during abstract task decomposition and scheduling to complete sub-tasks (\eg, the action of taking a photo requires a specific agent to press the shutter button). 
$\mathcal{C}_s$ denotes the need for multi-agents to avoid collisions—both agent-object and agent-agent—when sharing common spaces during task collaboration. 
$\mathcal{C}_t$ reflects the temporal-spatial sharing strategy among agents. Specifically, agents occupying a shared space at different time steps can improve collaboration efficiency (\eg, $a_2$ occupies a space in the $t_0$, while $a_4$ can occupies the same space in the $t_1$, achieving temporal-spatial sharing). 
This process can be expressed as follows:

{
\begin{equation}
\mathcal{G}^{\texttt{next}}, \mathcal{C}=\mathcal{F}_{\texttt{VLM}}(\mathcal{O}, \mathcal{G}_{\texttt{global}}, \mathcal{G}^{\texttt{pre}},f^{\texttt{pre}})
\end{equation}
}

\paragraph{RoboBrain} RoboBrain then generates trajectory sequences $traj_1, ..., traj_n$ for each agent to complete the corresponding subgoals by leveraging visual programming to invoke predefined motion primitives; these trajectories are unconstrained and may have potential failure risks. 
The detailed trajectory generation process is provided in the supplementary materials. 
Each agent executes its trajectory sequence along the time dimension, operating serially within the agent and in parallel between agents. 
This process is continuously monitored by RoboChecker to ensure the logical, spatial, and temporal validity of the subgoal trajectories.
%
The constraints generated by RoboBrain are confined to the textual space, making them ineffective for directly constraining trajectory data. 
Therefore, special interfaces are required to transform the textual constraints into specific representations that can directly interact with the real world and constrain the trajectory data.

\paragraph{RoboChecker} In RoboChecker, we provide GPT-4o~\cite{achiam2023gpt} with the textual constraints $\mathcal{C}$ for constraint-aware visual programming. 
For each constraint $c_i$ (\eg, ``Avoid collision between Agent\_2 and other Agents"), a corresponding interface $h_i$ which is define in Sec.~\ref{sec:method_constraint_interface} is created based on textual constraints $c_i$, addition with generated trajectory and other information needed (RGB images $\mathcal{O}$, robot states $\mathcal{S} = \{s_1,...s_n\}$).
Evaluation protocol is then generated (\ie, check code for trajectory), based on $h_i$ and the textual constraints $\mathcal{C}$.
This protocol evaluates whether the constraints are violated at the current time step by analyzing the interface $h_i$.
It returns a boolean indicating whether a constraint violation has occurred and a string describing the reason for the violation. 
If the protocol returns False, trajectory execution halts immediately, and the accompanying string is used as feedback ($f^{\texttt{pre}}$) for re-planning. Otherwise, the subgoal is marked as completed. In either case, the process repeats. Fully validated trajectories and observation sequences are served as part of the RoboFactory benchmark dataset.

\subsection{Constraint Interface}
\label{sec:method_constraint_interface}
To enable compositional constraints to govern decision-making in real-world scenarios, we need to model the abstract textual constraints into concrete criteria that can interact with the real world called \textbf{Constraint Interface}.
%
This ensures that RoboChecker can effectively restrict the set of generated trajectories. We have modeled the compositional constraints using four physical representation methods.

We model the logical constraint using two interaction logics in the physical world, namely interaction position and interaction direction, as shown in Fig.~\ref{fig:constraint}(a).
\textbf{Interaction Position}: For each 3D asset, we annotate the interaction positions. Different positions represent different interaction logics. For instance, grasping a camera and using a camera have distinct interaction positions.
\textbf{Interaction Direction}: Similarly, for every 3D asset, we mark the interaction directions. Different interaction behaviors between the robotic arm and the object follow different directional logics. For example, pressing the camera shutter requires the gripper of the robotic arm to move in the direct-facing direction of the shutter. In RoboChecker, based on the interaction form between the agent and the object, it will determine whether the current trajectory and grasping pose violate the logical constraints of interaction position and direction.

We analyze the current scene and establish a \textbf{3D occupancy} interface to implement spatial constraints, as shown in Fig.~\ref{fig:constraint}(b).
Specifically, we conduct depth estimation of the current 3D scene by utilizing hardware devices (such as depth cameras) or depth estimation methods~\cite{depthanything,depth_anything_v2}. Then, based on the current states of the robotic arm, we calculate the absolute coordinates of each joint point within the current space. By integrating the depth information, we obtain the occupancy information of the robotic arm and the objects. It is worth noting that we set a voxel with a size of 5cm*5cm*5cm as the basic discrete occupancy unit to reduce the computational cost. In RoboChecker, according to the occupancy relationship between the agent whose position is changing and other elements in the scene, it will determine whether a collision occurs and whether the spatial logic is violated.

 
We perform \textbf{dynamic occupancy modeling} for all intelligent agents that need to move under a sub-goal set $\mathcal{G}_{\texttt{next}}$ to account for temporal constraints, as shown in Fig.~\ref{fig:constraint}(c).
In cooperative tasks involving multiple embodied intelligent agents, the behaviors of robotic arms often occur simultaneously. Consequently, relying solely on spatial occupancy constraints can result in irrational scheduling, which significantly reduces task completion efficiency.
To address this, RoboChecker utilizes the temporal occupancy information of these intelligent agents to detect and prevent irrational scheduling as well as violations of temporal logic.

By establishing various interfaces required for compositional constraints, RoboChecker can convert abstract text-based constraints into representational forms that can interact with real-world decision-making. This enables the constraint of unrestricted trajectories, thereby generating safe and efficient multi-agent collaborative data.

 \begin{figure}[t]
  \centering
  \includegraphics[width=1\linewidth]{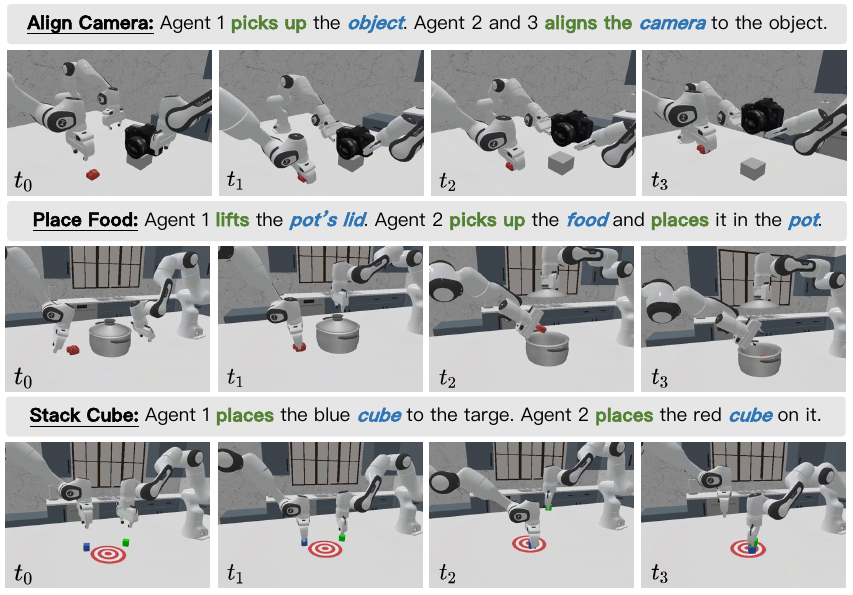}
  \caption{Demonstrations of the \mname{} Benchmark.}
  \vspace{-0.3cm}
\label{fig:method benchmark demos} 
\end{figure}

\subsection{Benchmark}
\label{sec:method_benchmark}
Based on the methods described above, we propose the \mname{} Benchmark, which is built on the ManiSkill simulator \cite{taomaniskill3}, an open-source platform for robot simulation. Tab.~\ref{tab:method benchmark comparison} demonstrates the comparison between other embodied benchmarks and the \mname{} benchmark. Our \mname{} benchmark features multi-agent tasks and the integration of high-level planning and low-level controlling. It includes 11 tasks across environments with varying numbers of agents, built based on the Franka Emika Panda Arm, a 7-DoF robotic manipulator equipped with an end-effector that enables flexible manipulation tasks, as depicted in Fig.~\ref{fig:method benchmark demos}. We utilize publicly available 3D assets from sources such as the PartNet-Mobility Dataset~\cite{xiang2020sapien}. For each task scenario, we configured an ego-centric camera for each agent and a global camera for all agents.

Our benchmark emphasizes efficient collaboration and coordination among agents in multi-agent environments. Agents must work together to complete specific tasks. For instance, in the task of Place Food, one agent must open the pot lid before another can place the food inside. 
The tasks involve diverse asset types, and the initial settings (\eg, asset placement) are randomly assigned using random seeds and can be easily replayed using the same seed. 
This domain randomization approach can effectively increase the diversity of training scenarios.
In \mname{} benchmark, 150 sets of data have been pre-collected for each task in form of camera RGB image observations, joint action of the robotic arms. More details can be found in the supplementary materials.

\begin{table}[t]
    \caption{Comparisons between \mname{} and other embodied benchmarks. It features multi-agent tasks and the integration of high-level planning and low-level controlling.}
    \centering
    \resizebox{1.\linewidth}{!}{
    \begin{tabular}{c|c|c|c}
        \toprule
       Benchmark  & Single-agent & Multi-agent & Task Level \\
       \hline
        EgoPlan-Bench~\cite{chen2023egoplan} & \cmark & \xmark & Plan \\
        MMWorld~\cite{hemmworld} & \cmark & \xmark & Plan \\
        VAB~\cite{liu2024visualagentbench} & \cmark & \xmark & Plan \\
        RoboCasa~\cite{soroush2024robocasa} & \cmark & \xmark & Plan \\
         RoboTwin~\cite{mu2024robotwin} & \cmark & \xmark & Plan \& Control \\
        RoboFactory(Ours) & \cmark & \cmark & Plan \& Control \\
        \bottomrule
    \end{tabular}
    }
    \vspace{-0.3cm}
    \label{tab:method benchmark comparison}
\end{table}

%% file: sec/4_exp.tex
\begin{table*}[t]
\centering
\caption{DP baseline performance results. We report the success rate across benchmark tasks with different amounts of demonstration data. }
\begin{tabular}{p{2cm}p{4.3cm}|p{2cm}p{2cm}p{2cm}}
\bottomrule[1pt]
\multirow{2}{*}{~~Task Level} & \multirow{2}{*}{~~~~~~~~~~~~~~~Task Name} & \multicolumn{3}{c}{~~Success Rate} \\

& & ~~~~50 Demo & ~~~~100 Demo  & ~~~~150 Demo\\
\hline
\multirow{4}{*}{\makecell[c]{~~~~1-Agent}} 
 &~~~~~~~~~~~~~~~~Pick Meat             &~~~~~~~~32\%  &~~~~~~~~\textbf{61\%}    &~~~~~~~~58\%\\
 &~~~~~~~~~~~~~~~Stack Cube             &~~~~~~~~17\%  &~~~~~~~~38\%   &~~~~~~~~\textbf{44\%}\\
 &~~~~~~~~~~~~~~~Strike Cube             &~~~~~~~~26\%  &~~~~~~~~42\%    &~~~~~~~~\textbf{45\%}\\
 &\cellcolor{gray!15}{~~~~~~~~~~~~~~~~~~Average}         & ~~~~~~~~\cellcolor{gray!15}{25\%} & ~~~~~~~~\cellcolor{gray!15}{47\%}   & ~~~~~~~~\cellcolor{gray!15}{\textbf{49\%}} \\
\hline
\multirow{5}{*}{\makecell[c]{~~~~2-Agent}}  
 &~~~~~~~~~~~~~~~~Pass Shoe             &~~~~~~~~~9\%  &~~~~~~~~\textbf{20\%}    &~~~~~~~~12\%\\
 &~~~~~~~~~~~~~~~~Place Food           &~~~~~~~~~5\%  &~~~~~~~~\textbf{23\%}    &~~~~~~~~20\%\\
 &~~~~~~~~~~~~~~~~Lift Barrier           &~~~~~~~~24\%  &~~~~~~~~\textbf{60\%}    &~~~~~~~~58\%\\
 &~~~~~~Two Robots Stack Cube     &~~~~~~~~14\%  &~~~~~~~~\textbf{27\%}    &~~~~~~~~20\%\\
 &\cellcolor{gray!15}{}{~~~~~~~~~~~~~~~~~~Average}         & ~~~~~~~~\cellcolor{gray!15}{13\%} & ~~~~~~~\cellcolor{gray!15}{\textbf{32.5\%}}  &~~~~~~~\cellcolor{gray!15}{27.5\%}\\
\hline
\multirow{3}{*}{\makecell[c]{~~~~3-Agent}}  
 &~~~~~~~~~~Camera Alignment            &~~~~~~~~~7\%  &~~~~~~~~10\%    &~~~~~~~~\textbf{19\%}\\
 &~~~~Three Robots Stack Cube            &~~~~~~~~~8\%  &~~~~~~~~~~2\%    &~~~~~~~~\textbf{22\%}\\
&\cellcolor{gray!15}{}{~~~~~~~~~~~~~~~~~~Average}         & ~~~~~~~~\cellcolor{gray!15}{7.5\%} & ~~~~~~~~~\cellcolor{gray!15}{6\%}    &~~~~~~~\cellcolor{gray!15}{\textbf{20.5\%}}\\
\hline
\multirow{3}{*}{~~~~4-Agent} 
 &~~~~~~~~~~~~~~~~Take Photo             &~~~~~~~~~5\%  &~~~~~~~~~8\%\   &~~~~~~~~\textbf{20\%}\\
 &~~~~~Long Pipeline Delivery   & ~~~~~~~~~0\% &  ~~~~~~~~~0\% & ~~~~~~~~~0\%\\
 &\cellcolor{gray!15}{}{~~~~~~~~~~~~~~~~~~Average}         & ~~~~~~~~\cellcolor{gray!15}{2.5\%} & ~~~~~~~~~\cellcolor{gray!15}{4\%}  &~~~~~~~~\cellcolor{gray!15}{\textbf{10\%}}\\
\bottomrule[1pt]
\end{tabular}
\label{tab:sup results for dp tasks}
\end{table*}

\section{Experiment}
Our experiments consist of following three parts: \textbf{(1)} the evaluation of Diffusion Policy on RoboFactory Benchmark (Sec.~\ref{sec:evaluation of benchmark}); \textbf{(2)} comparison of different architectural design of multi-agent systems based on imitation learning (Sec.~\ref{sec:mil}); \textbf{(3)} ablation studies on different constraints in RoboFactory data generation (Sec.~\ref{sec:ablation study}).

\subsection{Evaluation of \mname{} Benchmark}
\label{sec:evaluation of benchmark}
We evaluate Diffusion Policy (DP)~\cite{chi2023diffusion}, a generative method based on imitation learning, across 11 tasks in the \mname{} benchmark to justify the effectiveness of our method. 
For each agent, an individual policy is trained by taking ego-centric observations in RGB as input, without considering the states and actions of other agents.
We train the policies using 50, 100, and 150 expert demonstration data per task. More details for training strategies can be found in the supplementary materials.

\begin{figure}[t]
  \centering
  \vspace{-0.2cm}
  \includegraphics[width=1\linewidth]{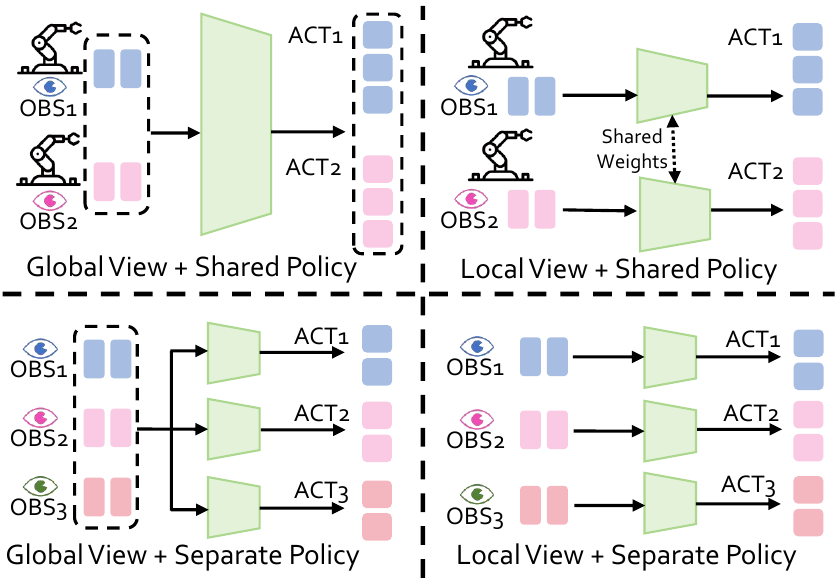}
  \caption{We design four multi-embodied agent imitation learning architectures. The \textbf{Global View} in the image input represents the observation containing all agents, and the \textbf{Local View} represents the ego-view observation of each agent. In policy training, \textbf{Shared Policy} indicates that all agents share a policy, and \textbf{Separate Policy} indicates that each agent trains an independent policy.
  }
\label{fig:arch} 
\end{figure}

Tab.~\ref{tab:sup results for dp tasks} demonstrates the main performance on \mname{} benchmark. 
First, the increase in success rates with additional training data emphasizes the necessity of RoboFactory in efficiently generating high-quality datasets. Specifically, tasks involving one, three, and four agents achieve optimal performance with 150 training demonstrations. These findings highlight the critical role of ample data, particularly in complex multi-agent environments.
At the same time, tasks involving two agents achieve optimal performance with only 100 demonstrations, which can be attributed to the relative simplicity of individual agent actions in these scenarios. Training with 150 demonstrations may have introduced overfitting, causing the model to learn unnecessary patterns that do not generalize well to testing environments. 
Besides, the decline in success rates with an increasing number of agents highlights the limitations of current methodologies. Specifically, tasks involving one, two, three, and four agents achieve average success rates of 49\%, 27.5\%, 20.5\%, and 10\%, respectively. This significant performance degradation in multi-agent tasks indicates the challenges in facilitating effective collaboration among multiple agents.
Moreover, the 0\% success rate in the task of Long Pipeline Delivery task highlights the shortcomings of diffusion policies in learning long-term temporal dependencies. These findings underscore significant opportunities for advancing imitation learning techniques in multi-agent systems to enhance performance.

\begin{table}[t]
\centering
\vspace{-0.2cm}
\caption{Results of four multi-embodied agent imitation learning architectures. We report the success rate on two tasks.}
\vspace{-0.2cm}
\resizebox{0.9\linewidth}{!}{
\begin{tabular}{cc|cc}
\bottomrule[1pt]
Policy & View Scope & Lift Barrier & Place Food \\
\hline
Shared & Global & 49\% & 5\% \\
Shared & Local & 4\% & 0\% \\
Separate & Global & 26\% & 17\% \\
Separate & Local & \textbf{58\%} & \textbf{20\%} \\
\bottomrule[1pt]
\end{tabular}
}
\label{tab:sup differ strategies}
\end{table}

\subsection{Multi-agent Imitation Learning}
\label{sec:mil}

In Sec.~\ref{sec:evaluation of benchmark}, we adapt the single-agent imitation learning framework to a multi-agent system, where each agent trains an independent policy based on its egocentric view. As illustrated in Figure \ref{fig:arch}, the architecture of multi-embodied agent imitation learning can be categorized into four types based on the observation space and policy sharing strategy:

\textbf{Global View and Shared Policy~(Arch1)}: All agents share the same global observation and use a single shared policy to produce a joint action sequence, which is then assigned to the corresponding agents.

\textbf{Local View and Shared Policy~(Arch2)}: Each agent has its own independent egoview observation, which is fed into the same shared policy (with shared parameters) to generate respective action sequences.

\textbf{Global View and Separate Policy~(Arch3)}: All agents share the same global observation, while each agent has its own separate policy (with unshared parameters) to generate individualized action sequences.

\textbf{Local View and Separate Policy~(Arch4)}: Each agent has its own independent egoview observation, and each agent uses its own separate policy (with unshared parameters) to generate individualized action sequences.

\begin{table}[t]
\centering
\caption{Ablation study for different constraints. We report the success rate(\%$\uparrow$) of effective data generation.}
\resizebox{1.\linewidth}{!}{
\begin{tabular}{ccc|cccc}
\bottomrule[1pt]
\multicolumn{3}{c|}{Components} & \multicolumn{3}{c}{Task Name} \\
\hline
\multirow{2}{*}{\makecell[c]{Logical}} & \multirow{2}{*}{\makecell[c]{Spatial}} & \multirow{2}{*}{\makecell[c]{Temporal}} & Lift & Three Robots & Take\\
& & & Barrier & Stack Cube & Photo\\
\hline
\cmark & \xmark & \xmark &80.2 & 62.5 &37.1\\
\cmark & \xmark & \cmark &85.4 & 84.2 &62.2\\
\cmark & \cmark & \xmark &95.2 & 92.7 &53.8\\
\cmark & \cmark & \cmark &\textbf{97.5} & \textbf{98.9}& \textbf{88.2}\\

\bottomrule[1pt]
\end{tabular}
}
\label{tab:sup ab succ rate}
\end{table}

We employ Arch4 as the pipeline to test all tasks in the RoboFactory benchmark. Additionally, we select two representative two-agent collaborative tasks—Lift Barrier and Food Place—to compare and analyze the four architectures. The experimental results are summarized in Table 1.
By comparing the first and second rows of Table 1, we observe that when a single shared policy needs to learn strategies for multiple agents from different ego-views, it must infer the agent ID currently executing an action and generate the corresponding action. This challenging setup causes the shared policy to struggle and leads to degraded performance~(49\%-5\%, 5\%-0\%).
By comparing the first two rows (shared policy) with the third and fourth rows (separate policy), we find that Separate Policy achieves better performance in the Food Place task~(5\%-17\%, 0\%-20\%). This improvement may be attributed to the Food Place task requiring distinct skills for the two robotic arms, where separate policies can specialize in learning the respective skills, thereby enhancing collaborative performance.
Finally, by comparing the third and fourth rows, we observe that using Local View under the Separate Policy setup improves task success rates~(26\%-58\%, 17\%-20\%). We hypothesize that this is because the egoview provides richer and more detailed information, enabling the policy to better handle fine-grained manipulations.

In summary, we design multiple multi-agent imitation learning architectures and conducted extensive experiments and analyses. We hope these findings can provide valuable insights for the future design of multi-agent imitation learning frameworks or multi-agent vision-language-action models, ultimately advancing the field of multi-embodied agent manipulation systems.

\subsection{Ablation Study}
\label{sec:ablation study}
Our ablation study aims to address two key questions: (1) Can the proposed constraints, particularly the compositional constraints, effectively improve the success rate of data generation (where higher success rates indicate faster data production)? (2) Do the proposed constraints lead to higher-quality data? We used the average episode length of the data as a metric to evaluate the quality. For tasks that are successfully executed, shorter data lengths suggest that agents can cooperate more effectively. In addition, shorter data lengths contribute to faster training and inference times. Tab.~\ref{tab:sup ab succ rate} and Tab.~\ref{tab:sup ab avg episode length} demonstrate the ablation studies on benchmark.

\vspace{-0.5mm} 
\paragraph{Spatial and temporal constraints significantly improve task success rates.}  The absence of spatial constraints leads to frequent robotic arm collisions during task execution, drastically reducing the success rate. Additionally, without spatial constraints, the robotic arms lack corrective spatial feedback, failing to adjust their positions properly. For temporal constraints, the decline in success rate mainly stems from two factors. First, tasks requiring simultaneous execution (\eg, Lift Barrier) fail due to improper synchronization. Second, errors in the execution order of robotic actions occur, such as incorrect stacking sequences in the Three Robot Stack Cube task.
\vspace{-1.2mm} 

\paragraph{Temporal constraints play an important role in enhancing data quality.}  Temporal constraints streamline task execution by identifying opportunities for simultaneous robotic arm operations, facilitating parallel execution, and reducing episode length in the dataset. Moreover, by analyzing and scheduling the sequence of robotic actions, they contribute to generating more structured and efficient data. When incorporated with occupancy grids, temporal constraints make fine-grained spatial awareness at each timestep, reducing erroneous failure feedback and fostering greater data diversity.  





%% file: sec/5_conclusion.tex
\begin{table}[t]
\centering
\caption{Ablation study for different constraints. We report the average episode length($\downarrow$) of the generated effective data.}
\resizebox{1.\linewidth}{!}{
\begin{tabular}{ccc|cccc}
\bottomrule[1pt]
\multicolumn{3}{c|}{Components} & \multicolumn{3}{c}{Task Name} \\
\hline
\multirow{2}{*}{\makecell[c]{Logical}} & \multirow{2}{*}{\makecell[c]{Spatial}} & \multirow{2}{*}{\makecell[c]{Temporal}} & Lift & Three Robots & Take \\
& & & Barrier & Stack Cube &  Photo\\
\hline
\cmark & \xmark & \xmark &123 &685& 407 \\
\cmark & \xmark & \cmark &92.8 &452 &238  \\
\cmark & \cmark & \xmark &115 &652 &325 \\
\cmark & \cmark & \cmark &\textbf{80.7} &\textbf{424} &\textbf{204} \\

\bottomrule[1pt]
\end{tabular}
}
\label{tab:sup ab avg episode length}
\vspace{-0.5mm}
\end{table}

\section{Conclusion}
\label{sec:conclusion}

We propose compositional constraints to tackle scalability challenges in transitioning from single-agent to multi-agent embodied systems and leverage compositional constraints to develop an automated data collection framework \mname{}. We introduce the first benchmark for embodied multi-agent manipulation. By deploying imitation learning and evaluating policy architectures on this benchmark, we systematically explore training strategies to advance safe and efficient multi-agent systems.

\vspace{-4.8mm}
\paragraph{Limitation} While \mname{} shows notable effectiveness, the constraints may struggle to accurately model intricate physical phenomena, potentially limiting their applicability in tasks requiring precise interactions.

%% file: sec/6_sup.tex
\appendix
\setcounter{page}{1}
\newpage
{\onecolumn
\centering
\Large
\textbf{\thetitle}\\
\vspace{0.5em}Supplementary Material \\
\vspace{1.0em}
}
\section{Data Generation}
In this section, we provide a detailed description of the process for effectively generating expert data. First, we elaborate on the details of RoboBrain, explaining how it generates the next subgoal and constraints. Next, we introduce the method for generating agent trajectories based on subgoals and constraints. Then, we describe the implementation of RoboChecker, an interface to integrate various constraints with data generation pipeline. Finally, we present tasks in the \mname{} benchmark along with its corresponding descriptions.
\vspace{-3mm} 
\paragraph{RoboBrain} In RoboBrain, we structure the following prompts for GPT-4o \cite{achiam2023gpt} to generate new subgoals based on the given information, such as task instructions, previous subgoals, and constraint violation feedback. Along with each subgoal, a set of constraints is generated, which can be categorized into three levels: logical, temporal, and spatial. These constraints are formulated as structured text to ensure that RoboChecker can accurately recognize the corresponding functions and verify whether the constraints are satisfied. The detailed description of prompts is as follows.

\begin{promptcode}[System Prompts]
    You are an AI system responsible for generating subgoals and constraints for a multi-agent robotic task. Your goal is to ensure that each agent receives a clearly defined subgoal while adhering to well-structured constraints. Constraints must be formatted correctly to enable validation and enforce coordination and collaboration among agents.
    
    Key Requirements
    - Generate at least one subgoal per agent based on the given task description.
    - Define explicit constraints for each agent, ensuring every constraint involves at least one agent.
    - Follow specific formatting rules to categorize constraints accurately.
    - Ensure all constraints are clear, actionable, and unambiguous to guide robotic agents effectively.
    
    Input Structure
    - Task Instruction: "{General description of the task}"
    - Global Observation: <image_global>
    - Agents Observation: [<image_1>,<image_2>....,<image_n>]
    - Previous Subgoals: "{Subgoals executed by each agent}"
    - Constraint Violation Feedback: "{List of feedback from violated constraints, if any}"
    
    Output Content
    - A set of subgoals and constraints based on the task requirements. Each constraint should follow these formatting rules according to its category:
    1. Logical Constraints:
       - Agent-specific condition: Agent-Specific Condition: Specifies a requirement for the behavior of a single agent.
           Example: "The gripper of Agent_1 must be perpendicular to {Object}."
       - Multi-agent condition: Defines a coordination rule between multiple agents.
           Example: "Agent_1 and Agent_2 must maintain a consistent gripper height.".
    2. Temporal Constraints:
       - Synchronization: Specifies whether agents can perform tasks simultaneously or share the same space.
           Example: "Agent_1 and Agent_3 perform tasks simultaneously without interference.".
       - Sequence: Defines the required order of actions between agents. 
           Example: "Agent_2 must complete the task before Agent_4 can begin their action.".
    3. Spatial Constraints:
       - Collision avoidance: Ensures agents do not interfere with each other or the environment.
           Example: "Agents must avoid colliding with each other when moving in close proximity.".
       - Space occupancy: Specifies spatial positioning rules to prevent conflicts.
           Example: "Agent_1 should not occupy the same space as Agent_3 in the designated area.".
    
    Output Example
    {
      "Subgoals": {
        "Agent_1": "{Clear and structured subgoals for Agent_1}",
        "Agent_2": "{Clear and structured subgoals for Agent_2}",
        ...
      },
      "Constraints": {
        "Logical": [
          {
            "Agent": "Agent_1",
            "Constraint": "The gripper of Agent_1 is perpendicular to {Object}."
          },
          {
            "Agents": ["Agent_2", "Agent_3"],
            "Constraint": "Keep the gripper height consistent between Agent_2 and Agent_3 to make the camera remain horizontal."
          }
        ],
        "Temporal": [
          {
            "Agents": ["Agent_2", "Agent_4"],
            "Constraint": Agent_2 and Agent_4 could share the same space chronologically."
          }
        ],
        "Spatial": [
          {
            "Agent": "Agent_2",
            "Constraint": "Avoid collision between Agent_2 and other Agents."
          },
          {
            "Agent": "Agent_4",
            "Constraint": "Avoid collision between Agent_4 and other Agents."
          }
        ]
      }
    }
\end{promptcode}

\vspace{-3mm} 
\paragraph{Trajectory Generation} Effectively converting these conceptual subgoals into precise robotic motion trajectories remains a significant challenge for large language models. Inspired by RoboTwin~\cite{mu2024robotwin}, we define a set of motion primitives, each represented as a Python function interface. By providing specific input parameters, these primitives generate corresponding motion trajectories. For instance, the MOVE primitive inputs an agent ID and a target position. Then, it computes a trajectory based on the current position of the robotic arm to generate the appropriate motion sequence. This approach allows large language models to focus on understanding the high-level logic of action interactions while avoiding direct involvement in low-level control signal computations.
\vspace{-2mm} 

\begin{figure}[t]
    \centering
    \includegraphics[width=\textwidth]{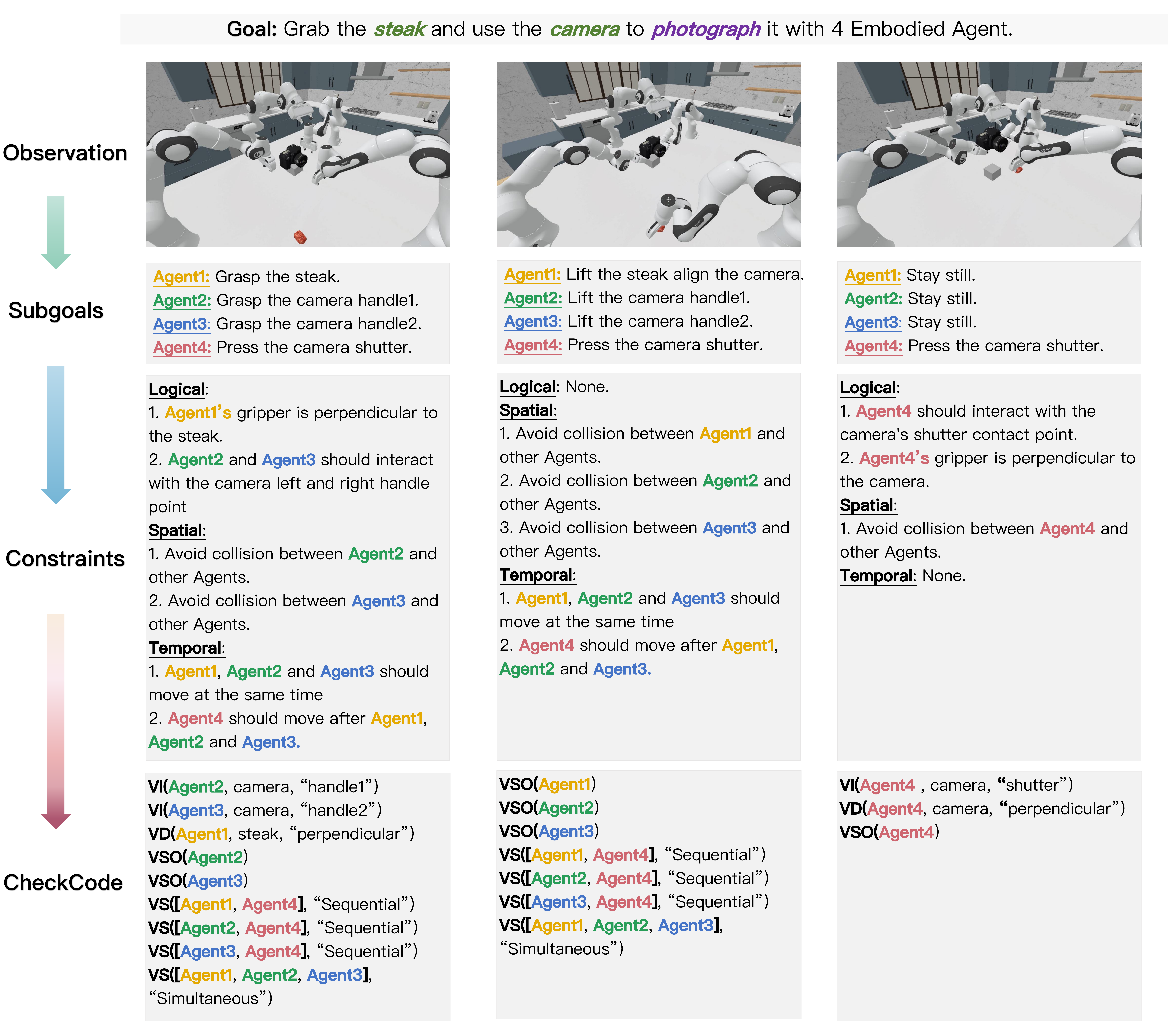}
    \caption{Demonstration of \textit{RoboChecker} is showcased in the complete execution of the Take Photo task. By analyzing constraints, \textit{RoboChecker} generates \textbf{CheckCode}, a composition of multiple interfaces. Specifically, VI stands for \textbf{Validate Interaction}, VD for \textbf{Validate Direction}, VSO for \textbf{Validate Spatial Occupancy}, and VS for \textbf{Validate Scheduling}. The CheckCode returns true only when all interfaces pass validation, indicating that the generated motion trajectory adheres to the compositional constraints. Otherwise, CheckCode identifies the failed interfaces and sends the feedback to \textit{RoboBrain}.}
    \label{fig:supp example of checkcode}
\end{figure}

\paragraph{RoboChecker} 
RoboChecker is designed to evaluate the validity and efficiency of generated motion trajectories, ensuring smooth execution while preventing collisions and inconsistencies. To achieve this, we define four key validation functions as optional interface type, each addressing a specific aspect of trajectory assessment: agent movement direction, interaction at contact points, spatial occupancy of trajectories, and the correctness of trajectory scheduling. The definitions of these functions are as follows:
\begin{itemize}
\item \textbf{Movement Direction Validation}: Ensures that the movement of each agent aligns with logical constraints. For instance, a robotic gripper should maintain an appropriate angle when grasping an object or adhere to a specific orientation during task execution to guarantee stable and effective interactions.
\item \textbf{Contact Point Interaction Validation}: Verifies whether an interaction with object or other agents meets expected conditions. In collaborative manipulation tasks, for example, multiple agents should grasp objects at appropriate positions to maintain stability during joint handling.
\item \textbf{Spatial Occupancy Validation}: Analyzes the spatial feasibility of an agent's trajectory, ensuring that it does not enter restricted zones or cause spatial conflicts. For instance, in confined environments, different agents' paths should remain non-overlapping to avoid collisions.
\item \textbf{Trajectory Scheduling Validation}: Assesses whether the execution order of motion trajectories adheres to temporal constraints. This includes ensuring that actions requiring synchronization occur simultaneously and that tasks with sequential dependencies are executed in the correct order. It will also analyze whether these operations can be executed simultaneously or follow a predefined sequence along the trajectory. For example, in a task where a lid must be opened before placing an object inside, the action of ``open lid" should precede the action of ``place object" in the trajectory plan.
\end{itemize}
\vspace{1mm}
These functions take two types of inputs: the current agent's trajectory and the constraints generated by RoboBrain. The constraints produced by RoboBrain are represented in textual form. To process these constraints effectively, we construct the following prompt to match each constraint with its corresponding function, extract the relevant parameters from the constraint text, and integrate them with the agent's trajectory for validation. In Fig~\ref{fig:supp example of checkcode}, we present the constraints of the Take Photo task along with the CheckCode generated through visual programming based on these constraint, which serves as the evaluation protocol, bridging textual constraints with the corresponding trajectory.

\begin{promptcode}[Validating Motion Trajectories with RoboChecker]
    You are an expert in robotic motion validation, responsible for ensuring that a given set of motion trajectories adheres to logical, spatial, and temporal constraints. Your task is to validate these trajectories based on predefined requirements, ensuring compliance with movement logic, spatial integrity, and execution order.
    To achieve accurate validation, you must:
    - Match each constraint to the appropriate validation function.
    - Extract relevant parameters from the constraint description.
    - Apply the corresponding validation rule to assess compliance.
    
    Validation Functions and Parameter Extraction
    Each constraint is assigned to a specific validation function, which extracts relevant parameters and applies the appropriate validation rule.
    1. Movement Direction Validation: Ensures that an agent maintains the required orientation during interactions.
        Extracted Parameters:
            Agent ID: The agent executing the movement.
            Target Object: The object involved in the interaction.
            Required Orientation: The necessary orientation for the gripper of the agent. 
        Formal Representation:
            (Agent_ID, Target_Object, Required_Orientation) -> Validate_Direction()
        Example Constraint:
            "The gripper of Agent_1 must be perpendicular to Object_A when grasping."
            (Agent_1, Object_A, perpendicular) -> Validate_Direction()
    2. Contact Point Interaction Validation: Ensures that agents interact with objects or other agents at the designated contact points.
        Extracted Parameters:
            Agent ID: The agent performing the interaction.
            Target Object: The Object involved in the interaction.
            Contact Point: The designated interaction point (\eg, left side of the object).
        Formal Representation:
            (Agent_ID, Target_Object, Contact_Point) -> Validate_Interaction()
        Example Constraint:
            "Agent_3 must grasp Object_B at its left point."
            (Agent_3, Object_B, left) -> Validate_Interaction()
    3. Spatial Occupancy Validation: Ensures that the movement of an agent does not result in spatial conflicts, such as entering restricted zones or colliding with other agents.
        Extracted Parameters:
            Agent IDs: The agents whose trajectories require validation.
        Formal Representation:
            (Agent_IDs) -> Validate_Spatial_Occupancy()
        Example Constraints and its parameters:
            "Agent_2 must not intersect with the trajectories of other agents."
            (Agent_2) -> Validate_Spatial_Occupancy()
    4. Trajectory Scheduling Validation: Ensures that the execution order of actions adheres to temporal constraints,  including:
        - Sequential dependencies, where one action must precede another.
        - Synchronized execution, where multiple agents must act simultaneously.
        Extracted Parameters:
            Agent IDs: The agents involved in the scheduling constraint.
            Task Dependency Type:
                - Sequential: Specifies an ordered execution sequence.
                - Simultaneous: Requires two agents to perform actions at the same time.
        Formal Representation:
            (Agent_IDs, Task_Dependency_Type) -> Validate_Scheduling()
        Example Constraints and its parameters:
            "Agent_4 must place Object_C only after Agent_5 opens the container."
            ([Agent_5, Agent_4], "Sequential") -> Validate_Scheduling()
\end{promptcode}
\vspace{-2mm} 

\paragraph{Tasks} Our benchmark dataset includes 11 tasks. Table \ref{tab:supple_task_description} presents the number of agents for each task, the task description, and the corresponding target condition. While single-agent tasks can assess the robotic arm's interaction capabilities, our primary focus is on multi-agent tasks, which evaluate the coordination and cooperation abilities between agents.

\section{Experimental Setup}
\subsection{Training Details}
We adopt the CNN-based Diffusion Policy as our base model, with a prediction horizon of 8, observation steps are set to 3, and action steps are set to 6. 
For the dataloader, we use a batch size of 128. 
The optimizer is set to \texttt{torch.optim.AdamW} with a learning rate of $1.0 \times 10^{-4}$, betas in the range of $[0.95, 0.999]$, and $\epsilon$ set to $1.0 \times 10^{-8}$. 
The learning rate warmup lasts 500 steps, and we train for 300 epochs for all tasks in the benchmark. 
The training process is conducted on a single Nvidia RTX 4090 GPU. For 150 demonstration samples with average episode length of 205, the training time is around 5 hours.

\paragraph{Different Training Strategies} Each Franka Emika Panda robotic arm consists of seven rotational joints, with an additional one-dimensional action for the gripper (as both left and right grippers maintain the same width), resulting in an action space of dimension 8 per agent. We design four different multi-agent DP strategy modes:
\begin{itemize}
    \item Global View and Shared Policy: All agents share a global observation that includes every agent in the environment. For a task with $N$ agents, their actions are concatenated to form a joint action of dimension $8N$. A single model is trained using the global view and the $8N$-dimensional joint action.
    \item Local View and Shared Policy: Each agent has an individual local observation centered on its own action. To prevent catastrophic forgetting during training, we randomly shuffle the training data of all agents before inputting them into a shared model. A single model is trained with multiple local observations and the corresponding agent's 8-dimensional action.
    \item Global View and Separate Policy: All agents share a global observation, ensuring that all agents are included within it. However, each agent trains its own model to determine actions independently. The training input consists of the global view and an individual agent’s 8-dimensional action.
    \item Local View and Separate Policy: Each agent has an individual local observation centered on its own action, with its own perspective prioritized while incorporating surrounding environmental information. Each agent trains a separate model using its local view and corresponding 8-dimensional action.
\end{itemize}
All global and local views used for training are RGB images with a resolution of 320×240. We select the fourth training strategies as the baselines for the benchmark experiments.

\subsection{Evaluation Details}
To ensure smooth robotic arm movements during simulations, we employ interpolating operation for action trajectories generated by the Diffusion Policy.
For each task, we conduct evaluations across 100 distinct scene configurations, varying initial object placements and environmental conditions. We introduce a maximum action step limit for each task for evaluation of success rates. A failure is determined if the task is not completed within this limit. To set a reasonable threshold, we perform a warm-up test among 20 samples to estimate the average number of steps required to complete the task. The maximum action step limit is set twice this average.
The success criteria for each task, including the target conditions, are detailed in Table \ref{tab:supple_task_description}.

\begin{figure}[t]
    \centering
    \includegraphics[width=\textwidth]{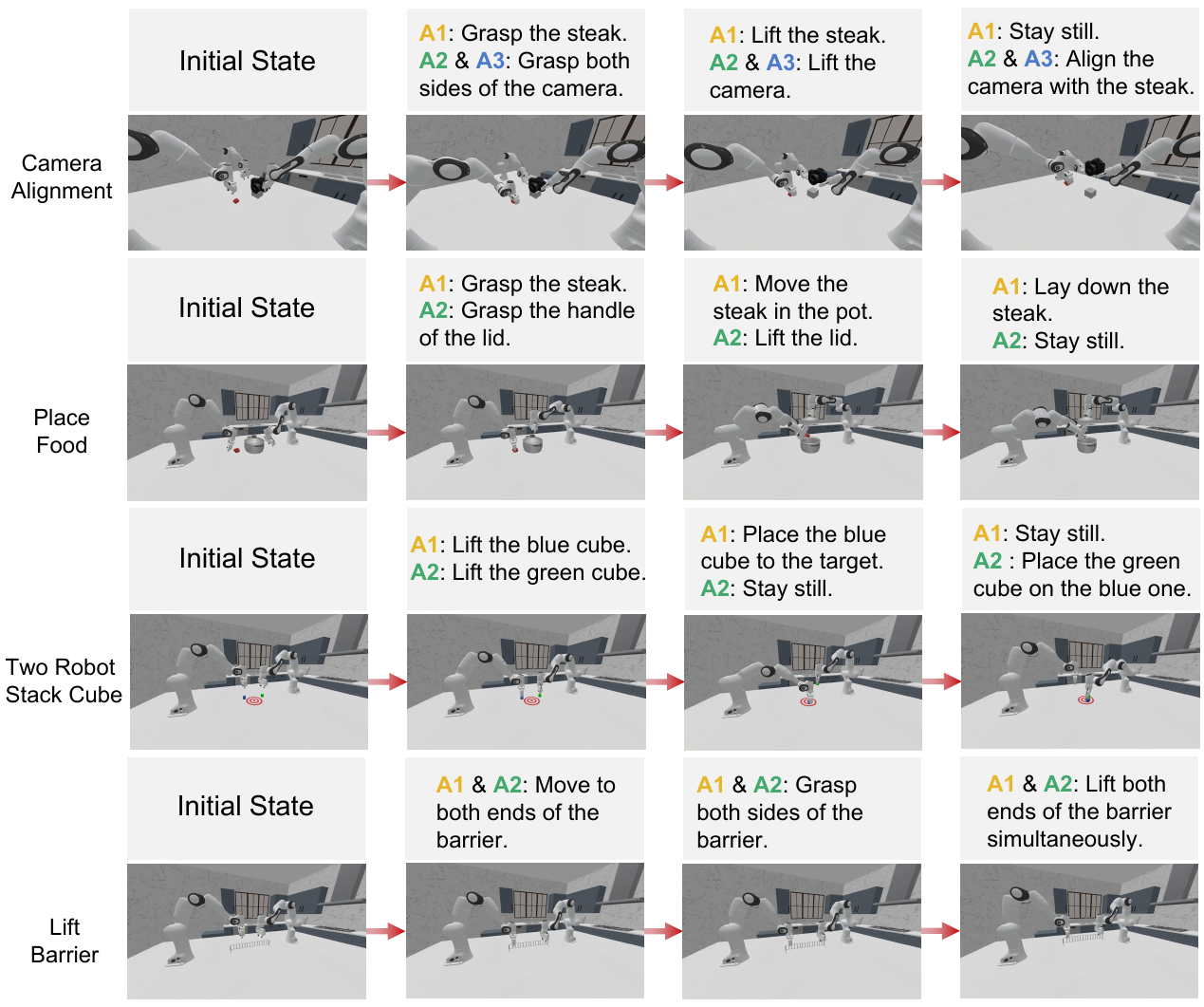}
    \caption{Demonstrations of tasks in the ~\mname{} Benchmark. For each task, the subgoals in each timestamp are displayed in the top row, and the observation is shown in the bottom row.
    }
    \label{fig:supp demos benchmark}
\end{figure}

\section{Demonstrations of Benchmark}
Fig.~\ref{fig:supp demos benchmark} demonstrates several tasks in the \mname{} benchmark. We visualize the observation of key timestamps with the corresponding subgoals generated from RoboBrain across 4 tasks (\textit{Camera Alignment}, \textit{Place Food}, \textit{Two Robot Stack Cube}, \textit{Lift Barrier}). 

\begin{table}[t]
    \scriptsize
    \centering
    \resizebox{1.\linewidth}{!}{
    \begin{tabular}{>{\centering}m{1.2cm}>{\centering} m{1cm} >{\scriptsize}m{4.1cm} >{\scriptsize}m{3.9cm}}
        Task & Agent Number &  \makecell{\centering Description} & \makecell{\centering Target Condition} \\
        \hline
        Pick Meat & 1 & There is a piece of meat placed on the table. A robotic arm picks up the meat and lifts it to a specified height. & The height of the meat reaches a predefined threshold. \\
        \hline
        Stack Cube & 1 & A blue cube and a red cube are placed on the table. A robotic arm picks up the blue cube and places it on top of the red cube. & The distance between the blue and red cubes is within a threshold, with the blue cube positioned at a greater height than the red cube.\\
        \hline
        Strike Cube & 1 & A hammer and a cube are placed on the table. A robotic arm first identifies an optimal grasping position to pick up the hammer, then moves it to a suitable position to strike the cube. & The hammerhead is positioned directly above the cube within a predefined distance threshold. \\
        \hline
        Lift Barrier & 2 & A long barrier is placed on the table. Two robotic arms simultaneously grasp both ends of the barrier and lift it to a specified height. & The barrier is elevated to the specified height while maintaining stability.\\
        \hline
        Pass Shoe & 2 & A shoe is placed on the table. One robotic arm grasps the shoe and passes it to the other one, which then places it at the target location. & The distance between the shoe and the target location is within a predefined threshold.\\ 
        \hline
        Place Food & 2 & A pot and a kind of food are placed on the table. One robotic arm lifts the pot's lid, while the other picks up the food and places it inside the pot. & The food is placed inside the pot, with the distance between the food and the center of the pot being within a predefined threshold.\\ 
        \hline
        Two Robots Stack Cube & 2 & A blue cube and a red cube are placed on the table. A robotic arm picks up the blue cube to a specified position, while the other places the red cube on top of it. & The blue cube is within the specified threshold distance from the target position. The distance between the blue and red cubes remains within a defined threshold, with the red cube positioned at a greater height than the blue cube.\\
        \hline
        Camera Alignment & 3 & A camera and an object are placed on the table. One robotic arm picks up the object to a specified position. The other two robotic arms grasp both sides of the camera and align it to the object. & The camera reaches a specified height, and the object is placed at the designated position that aligns with the camera. \\
        \hline
        Three Robots Stack Cube & 3 &  A blue cube, a red cube and a green cube are placed on the table. One robotic arm picks up the blue cube to a specified position. Another arm places the red cube on top of the blue one. The last arm places the green cube on top of the red one. & The blue cube is positioned within the specified target range. Additionally, the red cube is successfully placed on top of the blue cube, and the green cube is positioned atop the red cube. \\
        \hline
        Take Photo & 4 & A camera and an object are placed on the table. One robotic arm picks up the object and places it to a specified position. Another two robotic arms grasp both sides of the camera and align it to the object. The last robotic arm clicks the shutter. & The camera reaches a specified height, and the object is placed at the designated position that aligns with the camera. Additionally, the distance between the end effector of the last robotic arm and the camera's shutter is within a certain threshold.\\
        \hline
        Long Pipeline Delivery & 4 & A shoe is placed on the table. Three robotic arms grasp the shoe and pass it to the next robotic arm. The last robotic arm places the shoe to a specified position. & The distance between the shoe and the target location is within a predefined threshold.\\ 
        
    \end{tabular}
    }
    \caption{Task Descriptions for the \mname{} Benchmark}
    \label{tab:supple_task_description}
\end{table}


%% file: main.bbl
\begin{thebibliography}{52}
\providecommand{\natexlab}[1]{#1}
\providecommand{\url}[1]{\texttt{#1}}
\expandafter\ifx\csname urlstyle\endcsname\relax
  \providecommand{\doi}[1]{doi: #1}\else
  \providecommand{\doi}{doi: \begingroup \urlstyle{rm}\Url}\fi

\bibitem[Achiam et~al.(2023)Achiam, Adler, Agarwal, Ahmad, Akkaya, Aleman, Almeida, Altenschmidt, Altman, Anadkat, et~al.]{achiam2023gpt}
Josh Achiam, Steven Adler, Sandhini Agarwal, Lama Ahmad, Ilge Akkaya, Florencia~Leoni Aleman, Diogo Almeida, Janko Altenschmidt, Sam Altman, Shyamal Anadkat, et~al.
\newblock Gpt-4 technical report.
\newblock \emph{arXiv preprint arXiv:2303.08774}, 2023.

\bibitem[Black et~al.(2024)Black, Brown, Driess, Esmail, Equi, Finn, Fusai, Groom, Hausman, Ichter, et~al.]{black2024pi_0}
Kevin Black, Noah Brown, Danny Driess, Adnan Esmail, Michael Equi, Chelsea Finn, Niccolo Fusai, Lachy Groom, Karol Hausman, Brian Ichter, et~al.
\newblock pi0: A vision-language-action flow model for general robot control.
\newblock \emph{arXiv preprint arXiv:2410.24164}, 2024.

\bibitem[Buamanee et~al.(2024)Buamanee, Kobayashi, Uranishi, and Takemura]{buamanee2024bi}
Thanpimon Buamanee, Masato Kobayashi, Yuki Uranishi, and Haruo Takemura.
\newblock Bi-act: Bilateral control-based imitation learning via action chunking with transformer.
\newblock In \emph{2024 IEEE International Conference on Advanced Intelligent Mechatronics (AIM)}, pages 410--415. IEEE, 2024.

\bibitem[Chebotar et~al.(2023)Chebotar, Vuong, Hausman, Xia, Lu, Irpan, Kumar, Yu, Herzog, Pertsch, et~al.]{chebotar2023q}
Yevgen Chebotar, Quan Vuong, Karol Hausman, Fei Xia, Yao Lu, Alex Irpan, Aviral Kumar, Tianhe Yu, Alexander Herzog, Karl Pertsch, et~al.
\newblock Q-transformer: Scalable offline reinforcement learning via autoregressive q-functions.
\newblock In \emph{Conference on Robot Learning}, pages 3909--3928. PMLR, 2023.

\bibitem[Chen et~al.(2023)Chen, Ge, Ge, Ding, Li, Wang, Xu, Shan, and Liu]{chen2023egoplan}
Yi Chen, Yuying Ge, Yixiao Ge, Mingyu Ding, Bohao Li, Rui Wang, Ruifeng Xu, Ying Shan, and Xihui Liu.
\newblock Egoplan-bench: Benchmarking egocentric embodied planning with multimodal large language models.
\newblock \emph{CoRR}, 2023.

\bibitem[Chi et~al.(2023)Chi, Xu, Feng, Cousineau, Du, Burchfiel, Tedrake, and Song]{chi2023diffusion}
Cheng Chi, Zhenjia Xu, Siyuan Feng, Eric Cousineau, Yilun Du, Benjamin Burchfiel, Russ Tedrake, and Shuran Song.
\newblock Diffusion policy: Visuomotor policy learning via action diffusion.
\newblock \emph{The International Journal of Robotics Research}, page 02783649241273668, 2023.

\bibitem[Cho et~al.(2023)Cho, Zala, and Bansal]{Cho2023VPT2I}
Jaemin Cho, Abhay Zala, and Mohit Bansal.
\newblock Visual programming for text-to-image generation and evaluation.
\newblock In \emph{NeurIPS}, 2023.

\bibitem[Dalal et~al.(2023)Dalal, Mandlekar, Garrett, Handa, Salakhutdinov, and Fox]{Dalal2023ImitatingTA}
Murtaza Dalal, Ajay Mandlekar, Caelan~Reed Garrett, Ankur Handa, Ruslan Salakhutdinov, and Dieter Fox.
\newblock Imitating task and motion planning with visuomotor transformers.
\newblock In \emph{Conference on Robot Learning}, 2023.

\bibitem[Ebert et~al.(2021)Ebert, Yang, Schmeckpeper, Bucher, Georgakis, Daniilidis, Finn, and Levine]{ebert2021bridge}
Frederik Ebert, Yanlai Yang, Karl Schmeckpeper, Bernadette Bucher, Georgios Georgakis, Kostas Daniilidis, Chelsea Finn, and Sergey Levine.
\newblock Bridge data: Boosting generalization of robotic skills with cross-domain datasets.
\newblock \emph{arXiv preprint arXiv:2109.13396}, 2021.

\bibitem[Gupta and Kembhavi(2023)]{gupta2023visual}
Tanmay Gupta and Aniruddha Kembhavi.
\newblock Visual programming: Compositional visual reasoning without training.
\newblock In \emph{Proceedings of the IEEE/CVF Conference on Computer Vision and Pattern Recognition}, pages 14953--14962, 2023.

\bibitem[He et~al.()He, Feng, Zheng, Lu, Zhu, Li, Fan, Wang, Li, Yang, et~al.]{hemmworld}
Xuehai He, Weixi Feng, Kaizhi Zheng, Yujie Lu, Wanrong Zhu, Jiachen Li, Yue Fan, Jianfeng Wang, Linjie Li, Zhengyuan Yang, et~al.
\newblock Mmworld: Towards multi-discipline multi-faceted world model evaluation in videos.
\newblock In \emph{The Thirteenth International Conference on Learning Representations}.

\bibitem[Hong et~al.(2023)Hong, Zheng, Chen, Cheng, Wang, Zhang, Wang, Yau, Lin, Zhou, et~al.]{hong2023metagpt}
Sirui Hong, Xiawu Zheng, Jonathan Chen, Yuheng Cheng, Jinlin Wang, Ceyao Zhang, Zili Wang, Steven Ka~Shing Yau, Zijuan Lin, Liyang Zhou, et~al.
\newblock Metagpt: Meta programming for multi-agent collaborative framework.
\newblock \emph{arXiv preprint arXiv:2308.00352}, 3\penalty0 (4):\penalty0 6, 2023.

\bibitem[Huang et~al.(2024{\natexlab{a}})Huang, Li, Lam, Liang, Wang, Yuan, Jiao, Wang, Tu, and Lyu]{huang2024far}
Jen-tse Huang, Eric~John Li, Man~Ho Lam, Tian Liang, Wenxuan Wang, Youliang Yuan, Wenxiang Jiao, Xing Wang, Zhaopeng Tu, and Michael~R Lyu.
\newblock How far are we on the decision-making of llms? evaluating llms' gaming ability in multi-agent environments.
\newblock \emph{arXiv preprint arXiv:2403.11807}, 2024{\natexlab{a}}.

\bibitem[Huang et~al.(2024{\natexlab{b}})Huang, Qin, Lu, Wang, Huang, Shan, and Zhang]{huang2024story3d}
Yuzhou Huang, Yiran Qin, Shunlin Lu, Xintao Wang, Rui Huang, Ying Shan, and Ruimao Zhang.
\newblock Story3d-agent: Exploring 3d storytelling visualization with large language models.
\newblock \emph{arXiv preprint arXiv:2408.11801}, 2024{\natexlab{b}}.

\bibitem[James et~al.(2020)James, Ma, Arrojo, and Davison]{james2020rlbench}
Stephen James, Zicong Ma, David~Rovick Arrojo, and Andrew~J Davison.
\newblock Rlbench: The robot learning benchmark \& learning environment.
\newblock \emph{IEEE Robotics and Automation Letters}, 5\penalty0 (2):\penalty0 3019--3026, 2020.

\bibitem[Jang et~al.(2022)Jang, Irpan, Khansari, Kappler, Ebert, Lynch, Levine, and Finn]{jang2022bc}
Eric Jang, Alex Irpan, Mohi Khansari, Daniel Kappler, Frederik Ebert, Corey Lynch, Sergey Levine, and Chelsea Finn.
\newblock Bc-z: Zero-shot task generalization with robotic imitation learning.
\newblock In \emph{Conference on Robot Learning}, pages 991--1002. PMLR, 2022.

\bibitem[Jiang et~al.(2023)Jiang, Gupta, Zhang, Wang, Dou, Chen, Fei-Fei, Anandkumar, Zhu, and Fan]{jiang2023vima}
Yunfan Jiang, Agrim Gupta, Zichen Zhang, Guanzhi Wang, Yongqiang Dou, Yanjun Chen, Li Fei-Fei, Anima Anandkumar, Yuke Zhu, and Linxi Fan.
\newblock Vima: General robot manipulation with multimodal prompts.
\newblock In \emph{Fortieth International Conference on Machine Learning}, 2023.

\bibitem[Kalashnikov et~al.(2021)Kalashnikov, Varley, Chebotar, Swanson, Jonschkowski, Finn, Levine, and Hausman]{kalashnikov2021mt}
Dmitry Kalashnikov, Jacob Varley, Yevgen Chebotar, Benjamin Swanson, Rico Jonschkowski, Chelsea Finn, Sergey Levine, and Karol Hausman.
\newblock Mt-opt: Continuous multi-task robotic reinforcement learning at scale.
\newblock \emph{arXiv preprint arXiv:2104.08212}, 2021.

\bibitem[Kim et~al.(2024)Kim, Pertsch, Karamcheti, Xiao, Balakrishna, Nair, Rafailov, Foster, Lam, Sanketi, et~al.]{kim2024openvla}
Moo~Jin Kim, Karl Pertsch, Siddharth Karamcheti, Ted Xiao, Ashwin Balakrishna, Suraj Nair, Rafael Rafailov, Ethan Foster, Grace Lam, Pannag Sanketi, et~al.
\newblock Openvla: An open-source vision-language-action model.
\newblock \emph{arXiv preprint arXiv:2406.09246}, 2024.

\bibitem[Kumar et~al.(2022)Kumar, Singh, Ebert, Nakamoto, Yang, Finn, and Levine]{kumar2022pre}
Aviral Kumar, Anikait Singh, Frederik Ebert, Mitsuhiko Nakamoto, Yanlai Yang, Chelsea Finn, and Sergey Levine.
\newblock Pre-training for robots: Offline rl enables learning new tasks from a handful of trials.
\newblock \emph{arXiv preprint arXiv:2210.05178}, 2022.

\bibitem[Li et~al.(2023)Li, Hammoud, Itani, Khizbullin, and Ghanem]{li2023camel}
Guohao Li, Hasan Hammoud, Hani Itani, Dmitrii Khizbullin, and Bernard Ghanem.
\newblock Camel: Communicative agents for" mind" exploration of large language model society.
\newblock \emph{Advances in Neural Information Processing Systems}, 36:\penalty0 51991--52008, 2023.

\bibitem[Liang et~al.(2023)Liang, Huang, Xia, Xu, Hausman, Ichter, Florence, and Zeng]{liang2023code}
Jacky Liang, Wenlong Huang, Fei Xia, Peng Xu, Karol Hausman, Brian Ichter, Pete Florence, and Andy Zeng.
\newblock Code as policies: Language model programs for embodied control.
\newblock In \emph{2023 IEEE International Conference on Robotics and Automation (ICRA)}, pages 9493--9500. IEEE, 2023.

\bibitem[Liu et~al.(2024)Liu, Zhang, Gu, Iong, Xu, Song, Zhang, Lai, Liu, Zhao, et~al.]{liu2024visualagentbench}
Xiao Liu, Tianjie Zhang, Yu Gu, Iat~Long Iong, Yifan Xu, Xixuan Song, Shudan Zhang, Hanyu Lai, Xinyi Liu, Hanlin Zhao, et~al.
\newblock Visualagentbench: Towards large multimodal models as visual foundation agents.
\newblock \emph{arXiv preprint arXiv:2408.06327}, 2024.

\bibitem[Ma et~al.(2024)Ma, Zhou, Wang, Qiu, and Liang]{ma2024contrastive}
Teli Ma, Jiaming Zhou, Zifan Wang, Ronghe Qiu, and Junwei Liang.
\newblock Contrastive imitation learning for language-guided multi-task robotic manipulation.
\newblock \emph{arXiv preprint arXiv:2406.09738}, 2024.

\bibitem[Mandlekar et~al.(2018)Mandlekar, Zhu, Garg, Booher, Spero, Tung, Gao, Emmons, Gupta, Orbay, et~al.]{mandlekar2018roboturk}
Ajay Mandlekar, Yuke Zhu, Animesh Garg, Jonathan Booher, Max Spero, Albert Tung, Julian Gao, John Emmons, Anchit Gupta, Emre Orbay, et~al.
\newblock Roboturk: A crowdsourcing platform for robotic skill learning through imitation.
\newblock In \emph{Conference on Robot Learning}, pages 879--893. PMLR, 2018.

\bibitem[Mandlekar et~al.(2020)Mandlekar, Xu, Mart{\'\i}n-Mart{\'\i}n, Savarese, and Fei-Fei]{mandlekar2020learning}
Ajay Mandlekar, Danfei Xu, Roberto Mart{\'\i}n-Mart{\'\i}n, Silvio Savarese, and Li Fei-Fei.
\newblock Learning to generalize across long-horizon tasks from human demonstrations.
\newblock \emph{arXiv preprint arXiv:2003.06085}, 2020.

\bibitem[Mosquera et~al.(2024)Mosquera, Pinzon, Rios, Fonseca, Giraldo, Quijano, and Manrique]{mosquera2024can}
Manuel Mosquera, Juan~Sebastian Pinzon, Manuel Rios, Yesid Fonseca, Luis~Felipe Giraldo, Nicanor Quijano, and Ruben Manrique.
\newblock Can llm-augmented autonomous agents cooperate?, an evaluation of their cooperative capabilities through melting pot.
\newblock \emph{arXiv preprint arXiv:2403.11381}, 2024.

\bibitem[Mu et~al.(2024{\natexlab{a}})Mu, Chen, Zhang, Chen, Yu, Ge, Chen, Liang, Hu, Tao, et~al.]{mu2024robocodex}
Yao Mu, Junting Chen, Qinglong Zhang, Shoufa Chen, Qiaojun Yu, Chongjian Ge, Runjian Chen, Zhixuan Liang, Mengkang Hu, Chaofan Tao, et~al.
\newblock Robocodex: Multimodal code generation for robotic behavior synthesis.
\newblock \emph{arXiv preprint arXiv:2402.16117}, 2024{\natexlab{a}}.

\bibitem[Mu et~al.(2024{\natexlab{b}})Mu, Chen, Peng, Chen, Gao, Zou, Lin, Xie, and Luo]{mu2024robotwin}
Yao Mu, Tianxing Chen, Shijia Peng, Zanxin Chen, Zeyu Gao, Yude Zou, Lunkai Lin, Zhiqiang Xie, and Ping Luo.
\newblock Robotwin: Dual-arm robot benchmark with generative digital twins (early version).
\newblock \emph{arXiv preprint arXiv:2409.02920}, 2024{\natexlab{b}}.

\bibitem[Nambiar et~al.(2024)Nambiar, Jonsson, and Tarkian]{nambiar2024automation}
Sanjay Nambiar, Marie Jonsson, and Mehdi Tarkian.
\newblock Automation in unstructured production environments using isaac sim: A flexible framework for dynamic robot adaptability.
\newblock \emph{Procedia CIRP}, 130:\penalty0 837--846, 2024.

\bibitem[Nasiriany et~al.(2024)Nasiriany, Maddukuri, Zhang, Parikh, Lo, Joshi, Mandlekar, and Zhu]{soroush2024robocasa}
Soroush Nasiriany, Abhiram Maddukuri, Lance Zhang, Adeet Parikh, Aaron Lo, Abhishek Joshi, Ajay Mandlekar, and Yuke Zhu.
\newblock Robocasa: Large-scale simulation of everyday tasks for generalist robots.
\newblock In \emph{Robotics: Science and Systems}, 2024.

\bibitem[Qin et~al.(2024{\natexlab{a}})Qin, Shi, Yu, Wang, Zhou, Li, Yin, Liu, Sheng, Shao, et~al.]{qin2024worldsimbench}
Yiran Qin, Zhelun Shi, Jiwen Yu, Xijun Wang, Enshen Zhou, Lijun Li, Zhenfei Yin, Xihui Liu, Lu Sheng, Jing Shao, et~al.
\newblock Worldsimbench: Towards video generation models as world simulators.
\newblock \emph{arXiv preprint arXiv:2410.18072}, 2024{\natexlab{a}}.

\bibitem[Qin et~al.(2024{\natexlab{b}})Qin, Zhou, Liu, Yin, Sheng, Zhang, Qiao, and Shao]{qin2024mp5}
Yiran Qin, Enshen Zhou, Qichang Liu, Zhenfei Yin, Lu Sheng, Ruimao Zhang, Yu Qiao, and Jing Shao.
\newblock Mp5: A multi-modal open-ended embodied system in minecraft via active perception.
\newblock In \emph{Proceedings of the IEEE/CVF Conference on Computer Vision and Pattern Recognition}, pages 16307--16316, 2024{\natexlab{b}}.

\bibitem[Qin et~al.(2025)Qin, Sun, Hong, Wang, and Zhang]{qin2025navigatediff}
Yiran Qin, Ao Sun, Yuze Hong, Benyou Wang, and Ruimao Zhang.
\newblock Navigatediff: Visual predictors are zero-shot navigation assistants.
\newblock \emph{arXiv preprint arXiv:2502.13894}, 2025.

\bibitem[Tao et~al.(2024)Tao, Xiang, Shukla, Qin, Hinrichsen, Yuan, Bao, Lin, Liu, kai Chan, Gao, Li, Mu, Xiao, Gurha, Huang, Calandra, Chen, Luo, and Su]{taomaniskill3}
Stone Tao, Fanbo Xiang, Arth Shukla, Yuzhe Qin, Xander Hinrichsen, Xiaodi Yuan, Chen Bao, Xinsong Lin, Yulin Liu, Tse kai Chan, Yuan Gao, Xuanlin Li, Tongzhou Mu, Nan Xiao, Arnav Gurha, Zhiao Huang, Roberto Calandra, Rui Chen, Shan Luo, and Hao Su.
\newblock Maniskill3: Gpu parallelized robotics simulation and rendering for generalizable embodied ai.
\newblock \emph{arXiv preprint arXiv:2410.00425}, 2024.

\bibitem[Touvron et~al.(2023)Touvron, Lavril, Izacard, Martinet, Lachaux, Lacroix, Rozi{\`e}re, Goyal, Hambro, Azhar, et~al.]{touvron2023llama}
Hugo Touvron, Thibaut Lavril, Gautier Izacard, Xavier Martinet, Marie-Anne Lachaux, Timoth{\'e}e Lacroix, Baptiste Rozi{\`e}re, Naman Goyal, Eric Hambro, Faisal Azhar, et~al.
\newblock Llama: Open and efficient foundation language models.
\newblock \emph{arXiv preprint arXiv:2302.13971}, 2023.

\bibitem[Vaswani et~al.(2017)Vaswani, Shazeer, Parmar, Uszkoreit, Jones, Gomez, Kaiser, and Polosukhin]{vaswani2017attention}
Ashish Vaswani, Noam Shazeer, Niki Parmar, Jakob Uszkoreit, Llion Jones, Aidan~N Gomez, {\L}ukasz Kaiser, and Illia Polosukhin.
\newblock Attention is all you need.
\newblock \emph{Advances in neural information processing systems}, 30, 2017.

\bibitem[Vemprala et~al.(2024)Vemprala, Bonatti, Bucker, and Kapoor]{10500490}
Sai~H. Vemprala, Rogerio Bonatti, Arthur Bucker, and Ashish Kapoor.
\newblock Chatgpt for robotics: Design principles and model abilities.
\newblock \emph{IEEE Access}, 12:\penalty0 55682--55696, 2024.

\bibitem[Venuto et~al.(2024)Venuto, Islam, Klissarov, Precup, Yang, and Anand]{venuto2024code}
David Venuto, Sami~Nur Islam, Martin Klissarov, Doina Precup, Sherry Yang, and Ankit Anand.
\newblock Code as reward: Empowering reinforcement learning with vlms.
\newblock \emph{arXiv preprint arXiv:2402.04764}, 2024.

\bibitem[Wang et~al.(2024)Wang, Shen, Liu, and Xie]{wang2024sibyl}
Yulong Wang, Tianhao Shen, Lifeng Liu, and Jian Xie.
\newblock Sibyl: Simple yet effective agent framework for complex real-world reasoning.
\newblock \emph{arXiv preprint arXiv:2407.10718}, 2024.

\bibitem[Wu et~al.(2023)Wu, Bansal, Zhang, Wu, Li, Zhu, Jiang, Zhang, Zhang, Liu, et~al.]{wu2023autogen}
Qingyun Wu, Gagan Bansal, Jieyu Zhang, Yiran Wu, Beibin Li, Erkang Zhu, Li Jiang, Xiaoyun Zhang, Shaokun Zhang, Jiale Liu, et~al.
\newblock Autogen: Enabling next-gen llm applications via multi-agent conversation.
\newblock \emph{arXiv preprint arXiv:2308.08155}, 2023.

\bibitem[Xiang et~al.(2020)Xiang, Qin, Mo, Xia, Zhu, Liu, Liu, Jiang, Yuan, Wang, et~al.]{xiang2020sapien}
Fanbo Xiang, Yuzhe Qin, Kaichun Mo, Yikuan Xia, Hao Zhu, Fangchen Liu, Minghua Liu, Hanxiao Jiang, Yifu Yuan, He Wang, et~al.
\newblock Sapien: A simulated part-based interactive environment.
\newblock In \emph{Proceedings of the IEEE/CVF conference on computer vision and pattern recognition}, pages 11097--11107, 2020.

\bibitem[Yang et~al.(2024{\natexlab{a}})Yang, Kang, Huang, Xu, Feng, and Zhao]{depthanything}
Lihe Yang, Bingyi Kang, Zilong Huang, Xiaogang Xu, Jiashi Feng, and Hengshuang Zhao.
\newblock Depth anything: Unleashing the power of large-scale unlabeled data.
\newblock In \emph{CVPR}, 2024{\natexlab{a}}.

\bibitem[Yang et~al.(2024{\natexlab{b}})Yang, Kang, Huang, Zhao, Xu, Feng, and Zhao]{depth_anything_v2}
Lihe Yang, Bingyi Kang, Zilong Huang, Zhen Zhao, Xiaogang Xu, Jiashi Feng, and Hengshuang Zhao.
\newblock Depth anything v2.
\newblock \emph{arXiv:2406.09414}, 2024{\natexlab{b}}.

\bibitem[Yang et~al.(2024{\natexlab{c}})Yang, Zhang, Zheng, Jiang, Gan, Wang, Ling, Chen, Ma, Dong, et~al.]{yang2024oasis}
Ziyi Yang, Zaibin Zhang, Zirui Zheng, Yuxian Jiang, Ziyue Gan, Zhiyu Wang, Zijian Ling, Jinsong Chen, Martz Ma, Bowen Dong, et~al.
\newblock Oasis: Open agents social interaction simulations on one million agents.
\newblock \emph{arXiv preprint arXiv:2411.11581}, 2024{\natexlab{c}}.

\bibitem[Yu et~al.(2024)Yu, Fu, Deng, and Han]{yu2024mineland}
Xianhao Yu, Jiaqi Fu, Renjia Deng, and Wenjuan Han.
\newblock Mineland: Simulating large-scale multi-agent interactions with limited multimodal senses and physical needs.
\newblock \emph{arXiv preprint arXiv:2403.19267}, 2024.

\bibitem[Ze et~al.(2024)Ze, Zhang, Zhang, Hu, Wang, and Xu]{Ze2024DP3}
Yanjie Ze, Gu Zhang, Kangning Zhang, Chenyuan Hu, Muhan Wang, and Huazhe Xu.
\newblock 3d diffusion policy: Generalizable visuomotor policy learning via simple 3d representations.
\newblock In \emph{Proceedings of Robotics: Science and Systems (RSS)}, 2024.

\bibitem[Zhao et~al.(2024)Zhao, Huang, Xu, Lin, Liu, and Huang]{zhao2024expel}
Andrew Zhao, Daniel Huang, Quentin Xu, Matthieu Lin, Yong-Jin Liu, and Gao Huang.
\newblock Expel: Llm agents are experiential learners.
\newblock In \emph{Proceedings of the AAAI Conference on Artificial Intelligence}, pages 19632--19642, 2024.

\bibitem[Zhao et~al.(2023)Zhao, Kumar, Levine, and Finn]{zhao2023learning}
Tony~Z Zhao, Vikash Kumar, Sergey Levine, and Chelsea Finn.
\newblock Learning fine-grained bimanual manipulation with low-cost hardware.
\newblock \emph{arXiv preprint arXiv:2304.13705}, 2023.

\bibitem[Zhou et~al.(2024{\natexlab{a}})Zhou, Qin, Yin, Huang, Zhang, Sheng, Qiao, and Shao]{zhou2024minedreamer}
Enshen Zhou, Yiran Qin, Zhenfei Yin, Yuzhou Huang, Ruimao Zhang, Lu Sheng, Yu Qiao, and Jing Shao.
\newblock Minedreamer: Learning to follow instructions via chain-of-imagination for simulated-world control.
\newblock \emph{arXiv preprint arXiv:2403.12037}, 2024{\natexlab{a}}.

\bibitem[Zhou et~al.(2024{\natexlab{b}})Zhou, Su, Chi, Zhang, Wang, Huang, Sheng, and Wang]{zhou2024code}
Enshen Zhou, Qi Su, Cheng Chi, Zhizheng Zhang, Zhongyuan Wang, Tiejun Huang, Lu Sheng, and He Wang.
\newblock Code-as-monitor: Constraint-aware visual programming for reactive and proactive robotic failure detection.
\newblock \emph{arXiv preprint arXiv:2412.04455}, 2024{\natexlab{b}}.

\bibitem[Zhou et~al.(2023)Zhou, Xu, Zhu, Zhou, Lo, Sridhar, Cheng, Ou, Bisk, Fried, et~al.]{zhou2023webarena}
Shuyan Zhou, Frank~F Xu, Hao Zhu, Xuhui Zhou, Robert Lo, Abishek Sridhar, Xianyi Cheng, Tianyue Ou, Yonatan Bisk, Daniel Fried, et~al.
\newblock Webarena: A realistic web environment for building autonomous agents.
\newblock \emph{arXiv preprint arXiv:2307.13854}, 2023.

\end{thebibliography}
